\documentclass[final]{cvpr}

\usepackage{lipsum}
\usepackage{amsthm}
\usepackage{times}
\usepackage{epsfig}
\usepackage{graphicx}
\usepackage{microtype}
\usepackage{graphicx}
\usepackage{subfigure}
\usepackage{booktabs} % for professional tables
\usepackage{amsmath}
\usepackage{amssymb}
\usepackage{diagbox}
\usepackage{url}
\usepackage{caption}
\usepackage{wrapfig}
\usepackage[super]{nth}
\usepackage{physics}
\usepackage{pifont}
\usepackage{amsmath}
\usepackage{relsize}
\usepackage{adjustbox}
\usepackage{enumitem}
\usepackage{algorithm}
\usepackage[noend]{algorithmic}
\usepackage{graphicx}
\usepackage{epstopdf}

\newtheorem{prop}{Proposition}

\newtheorem{defn}{Definition}

\usepackage{mathtools}

\DeclarePairedDelimiter\floor{\lfloor}{\rfloor}

\DeclareMathOperator*{\argmin}{arg\,min}

\makeatletter
\@namedef{ver@everyshi.sty}{}
\makeatother
\usepackage{tikz}

\newcommand{\cmark}{\ding{51}}%
\newcommand{\xmark}{\ding{55}}%
\newcommand\rom[1]{{\color{blue}[Romain: #1]}}

\usepackage{cite}
\usepackage[pagebackref=true,breaklinks=true,colorlinks,bookmarks=false]{hyperref}

\def\cvprPaperID{7840} % *** Enter the CVPR Paper ID here
\def\confYear{CVPR 2021}

\begin{document}

%%%%%%%%% TITLE
\title{Interpretable Image Clustering via Diffeomorphism-Aware $K$-Means}

\author{\textbf{Romain Cosentino}\\
Rice University\\
% For a paper whose authors are all at the same institution,
% omit the following lines up until the closing ``}''.
% Additional authors and addresses can be added with ``\and'',
% just like the second author.
% To save space, use either the email address or home page, not both
\and
\textbf{Randall Balestriero}\\
Rice University\\
\and
\textbf{Yanis Bahroun}\\
Flatiron Institute\\
\and
\textbf{Anirvan Sengupta}\\
Flatiron Institute \& Rutgers University \\
\and
\textbf{Richard Baraniuk}\\
Rice University\\
\and
\textbf{Behnaam Aazhang}\\
Rice University\\
}
    \vspace*{-5pt}
    {\let\newpage\relax\maketitle}

\vspace{-.3cm}
%%%%%%%%% ABSTRACT
\begin{abstract}
%We address the problem of 
We design an interpretable clustering algorithm aware of the nonlinear structure of image manifolds. Our approach leverages the interpretability of $K$-means applied in the image space while addressing its clustering performance issues. Specifically, we develop a measure of similarity between images and centroids that encompasses a general class of deformations: diffeomorphisms, rendering the clustering invariant to them. Our work leverages the Thin-Plate Spline interpolation technique to efficiently learn diffeomorphisms best characterizing the image manifolds. Extensive numerical simulations show that our approach competes with state-of-the-art methods on various datasets. 
\end{abstract}
%%%%%%%%% BODY TEXT
\vspace{-.3cm}
\section{Introduction}

\iffalse
With the number of recording devices increasing in every field of study, from biological recordings to digital advertisement, it has become essential to develop machine learning algorithms capable of analyzing complex interactions present in data. Among the various existing machine learning approaches, supervised learning remains the most commonly used. However, the cost and need for an expert to provide the supervision are substantial limitations. As a result, unsupervised learning algorithms, which are less dependent on expert knowledge, remain crucial for discovering patterns in big data \cite{dolnicar2003using,xu2010clustering}. A specific branch of unsupervised learning central to our work is {\em clustering}. Clustering algorithms aim to organize data into groups based on distinct similarities and are often considered a bridge between supervised and unsupervised learning.
\fi
With the amount of observations increasing in every field, from biological experiments to digital advertisement, it has become essential to develop machine learning algorithms capable of analyzing subtle structures present in data. Despite recent giant strides in supervised learning, the cost and need for an expert to provide the supervision remain substantial limitations. Thus, unsupervised learning algorithms, which are less dependent on expert knowledge, remain crucial for discovering patterns in big data \cite{dolnicar2003using,xu2010clustering}. A specific branch of unsupervised learning central to our work is {\em clustering}. Clustering algorithms aim to organize data into groups based on distinct similarities and are often considered a bridge between supervised and unsupervised learning. It is important that clustering be interpretable both for data exploration and for deriving insights \cite{bertsimas2020interpretable,greene2005producing}.

\iffalse
An integral part of the design of clustering algorithms is the choice of an appropriate metric space \cite{he2013k,frey2002fast,990920}. The clustering of data points requires measuring the distance between each of them and the centroids for the correct assignment to a cluster. In particular, the famous $K$-means clustering algorithm
\cite{macqueen1967some}, known for its simplicity, efficiency, and interpretability, relies on the Euclidean distance. However, when dealing with metric spaces, one often has to choose between computational efficiency and the explanatory power of the relationship that is extracted. For example, the Euclidean distance makes the design of efficient algorithms easy, but it often leads to the extraction of trivial relationships between data points.
%
Besides, defining a distance between images lying in high-dimensional spaces has proved to be a critical bottleneck toward high-performance clustering algorithms \cite{steinbach2004challenges}.
\fi

The $K$-means clustering algorithm
\cite{macqueen1967some} is well-known for its simplicity, efficiency, and, in particular, interpretability. An integral part of the design of such a clustering algorithms is the choice of an appropriate metric space \cite{he2013k,frey2002fast,990920}. The clustering of data points requires measuring the distance between each of them and the centroids for the correct assignment to a cluster. The commonly used $K$-means algorithm relies on the Euclidean distance. However, when dealing with metric spaces, one often has to choose between computational efficiency and the explanatory power of the relationship that is extracted. While the Euclidean distance makes the design of efficient algorithms easy, this measure of similarity might miss subtle relationships between data points that are not superficially close by. Defining an appropriate measure of similarity between images lying in high-dimensional spaces is  a critical bottleneck to high-performance clustering algorithms \cite{steinbach2004challenges}.

\iffalse
A recent way to extract nonlinear relationships between data is built upon the concept of invariance \cite{mallat2016understanding,bruna2013invariant,oyallon2017scaling,cohen2019gauge}. In this work, we are particularly interested in the invariances induced by a metric, which is defined as the immutable reaction of a metric through the action of a transformation on the data. In computer vision, the study of affine invariant distance has a long tradition, initially inspired by neuroscience observations. In the early 1940s, the pioneering work in \cite{pitts1947we} suggested that the multi-layer network underlying our vision acts as a similarity detector that enables the detection of classes of equivalences and is still being actively investigated \cite{kriegeskorte2008matching,sengupta2018manifold}.
\fi

A recent way to extract nonlinear relationships between data is built upon the concept of invariance \cite{mallat2016understanding,bruna2013invariant,oyallon2017scaling,cohen2019gauge,anden2014deep}. In this work, we are particularly interested in measures of similarity that incorporate the action of a transformation on the data into its very definition. In computer vision, the study of affine invariant distance has a long tradition, initially inspired by neuroscience observations. In the early 1940s, the pioneering work in \cite{pitts1947we} suggested that the multi-layer network underlying our vision acts as a similarity detector that enables the detection of classes of equivalences and is still being actively investigated \cite{kriegeskorte2008matching,sengupta2018manifold}.

These findings have later been leveraged to provide powerful unsupervised computer vision algorithms \cite{kendall2009shape,werman1995similarity,berthilsson1998statistical}. When applied to images, these pioneering techniques usually require each image to be landmarked, i.e., to have annotated points on their most essential features \cite{begelfor2006affine}. Another critical set of work, e.g, the works in \cite{Sifre_2013_CVPR,czaja2018scattering,charalampidis2005modified,bruna2013invariant,le2002unsupervised}, focused on visual appearance instead of shape. They proposed the development of features that are invariant to specific transformations. However, it was noted that extracting these features reliably and consistently is not easy \cite{lim2004image}.
In the same vein, the use of the Dynamic Time Warping algorithm on flattened images has been used to understand and compare images being the results of transformations \cite{rath2003word,698671,santosh2010use}.

%Finally, t
The approach that has most inspired our work considers images as points in a high-dimensional space. In this space, similarities in appearance between images are quantified by their geometric proximity \cite{262951}. These methods are referred to as appearance manifold-based approaches and have been successfully used to cluster images \cite{basri1998clustering,su2001modified,1211332}.
Notable work \cite{fitzgibbon2002affine,simard2012transformation,lim2004image} has introduced several distances, which followed the development of metrics capturing symmetries in the data. A key observation of the authors was that the Euclidean distance between two images of faces of different individuals but with the same pose is always smaller than the Euclidean distance between two images of the same individual's faces in different poses. From these observations, in \cite{fitzgibbon2002affine,simard2012transformation,lim2004image}, the authors clearly expressed the need for distances to distinguish data based on their taxonomy and be invariant to their orientations, position in the images, and illumination conditions. Whereas the approaches mentioned above appear general, they are limited to the affine case, when in practice, images like in the example aforementioned are subject to more intricate and non-rigid symmetries. 

In this work, we show that it is essential to consider a broader class of transformations, diffeomorphisms, to capture the complex transformations an image can undertake while preserving its original identity. We thus develop a similarity measure that is invariant to this general class of deformation. As a result, by combining the popular $K$-means with our measure, we introduce the Deformation Invariant $K$-means (DI $K$-means) algorithm. Our approach is displaying improved performance on benchmarks, outperforming the Affine Invariant $K$-means as well as competing with state-of-the-art models.

Our contributions can be summarized as follows:
\begin{itemize}[leftmargin=*]
  \setlength{\itemsep}{0pt}
    \item We define an efficient parametrization of image transformations with respect to diffeomorphisms, Sec.~\ref{sec:background}.
    \item We introduce the {\em Deformation Invariant $K$-means algorithm} that addresses the limitations of the Affine Invariant $K$-means approach, as well as a novel centroid update rule that enables the learnability of centroid lying on the image manifold, Sec.~\ref{sec:RAI-Kmeans}.
    \item Finally, we show numerically that our unsupervised algorithm competes with state-of-the-art methods on various datasets while benefiting from interpretable results, Sec.~\ref{sec:experiments}.
\end{itemize}

\section{Image Transformations}
\label{sec:background}

In this work, we build a measure that enables us to capture the symmetries in the data. To do so, we focus our interest on the action of groups of transformation on the data. By action of a group on an image, we refer to the deformation of this image with respect to a particular transformation induced by a group. For instance, the rotation, translation, and shearing of an image can be considered the result of the action of elements of the affine group. An introduction to group transformations can be found in \cite{hall2015lie}.

In this section, we present two popular groups of transformation in computer vision, the affine and the diffeomorphism group, denoted by $\text{Aff}(\mathbb{R}^2)$ and $\text{Diff}(\mathbb{R}^2)$, respectively. 

%In this work, we are interested in building a metric that enables us to capture the symmetries in the data. To do so, we will focus our interest in the action group on the data. By action of a group on an image, we consider the deformation of this image with respect to a particular transformation induced by a group. 
%
%In particular, we will be interested in two well-known groups in computer vision, the affine group denoted by $\text{Aff}(\mathbb{R}^2)$ and the diffeomorphism group, $\text{Diff}(\mathbb{R}^2)$. 
%

\subsection{Transformation Groups}

\begin{figure}[!h]
   \centering
   \begin{minipage}{.32\linewidth}
   \includegraphics[width=\textwidth]{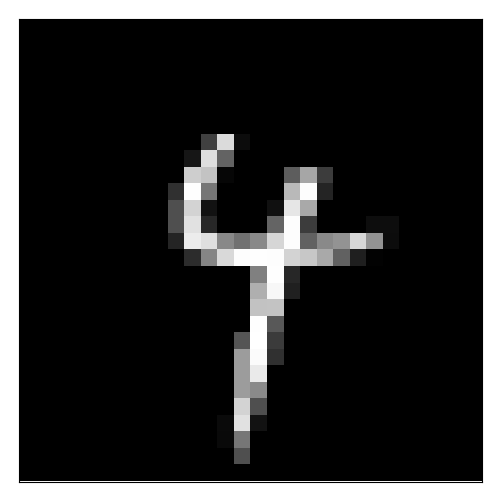}
   \end{minipage}
     \begin{minipage}{.32\linewidth}
   \includegraphics[width=\textwidth]{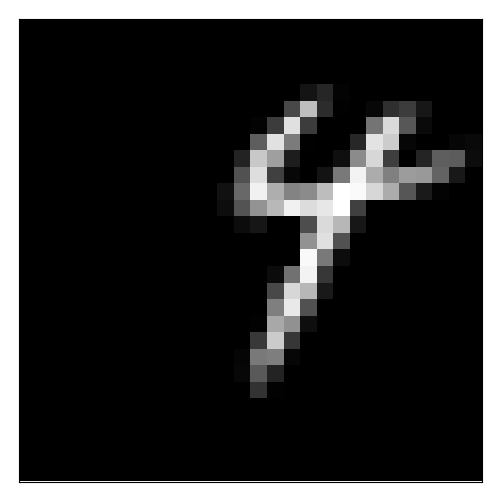}
   \end{minipage}
         \begin{minipage}{.32\linewidth}
   \includegraphics[width=\textwidth]{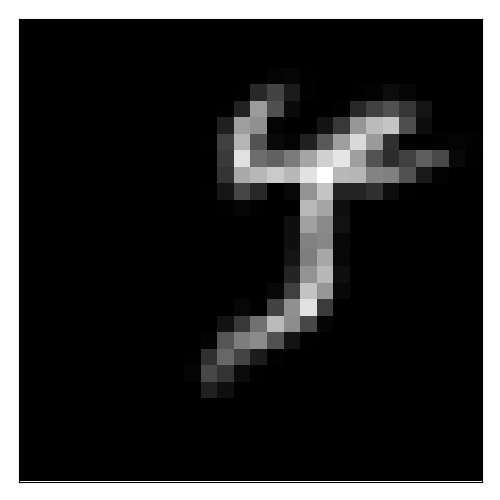}
   \end{minipage}
         \begin{minipage}{.32\linewidth}
   \includegraphics[width=\textwidth]{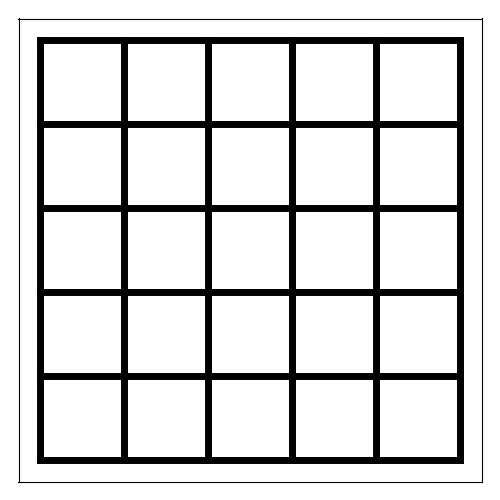}
   \end{minipage}
      \begin{minipage}{.32\linewidth}
   \includegraphics[width=\textwidth]{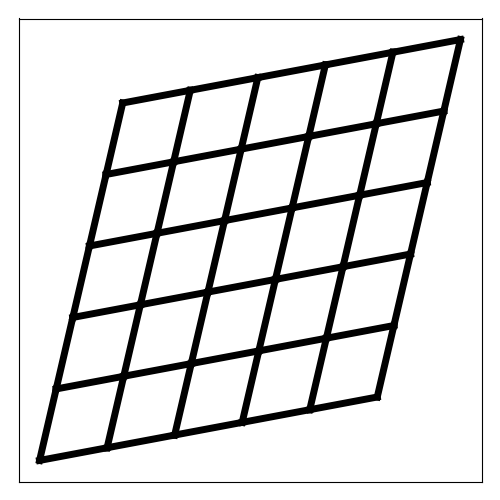}
   \end{minipage}       
      \begin{minipage}{.32\linewidth}
   \includegraphics[width=\textwidth]{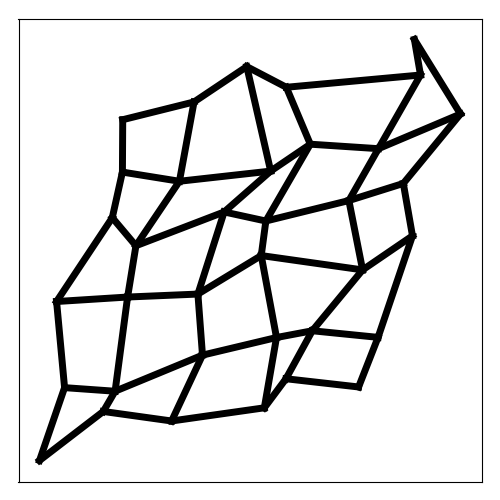}
   \end{minipage}   
   \caption{\textbf{Image Transformations - } Visualizations of a sample taken from the MNIST dataset and its transformed versions. Each image results from the application of the transformation induced by the grid displayed below it. (\textit{Left}) we observe the original image and its associated original transformation grid, which corresponds to the identity transform. (\textit{Middle)} the image has been transformed by the affine transformation induced by the associated grid. (\textit{Right}) the image transformed by the diffeomorphism using the TPS induced by the grid below it. Note that the associated grid on the bottom right is a diffeomorphism generalizing affine transformations.} %the affine transformations and the
   \label{fig:orig_to_diffeo}
 \end{figure}

The groups of transformation that we consider is applied to the coordinates of the pixels of an image, characterized by the real plane, $\mathbb{R}^2$. For example, in Fig.~\ref{fig:orig_to_diffeo}, the image is characterized by the intensity of the pixels located on a grid of $\mathbb{R}^2$. A transformation then acts on an image by changing the coordinates of its pixels in the plane, as shown by the deformed grids in Fig.~\ref{fig:orig_to_diffeo} .
Then given a transformation of the pixel coordinates, represented by a grid, the transformed image is rendered by bilinear interpolation as in Fig.~\ref{fig:orig_to_diffeo}. 
%depicted for a depiction of our transformation framework

Unlike previous work, we not only exploit the affine group, $\text{Aff}(\mathbb{R}^2)$, consisting of only $6$ parameters, but also the group of diffeomorphisms, $\text{Diff}(\mathbb{R}^2)$. The elements of $\text{Diff}(\mathbb{R}^2)$ are smooth and possibly nonlinear functions acting on $\mathbb{R}^2$, again representing the coordinates of the pixels of an image. In Fig.~\ref{fig:orig_to_diffeo}, we show an example of a diffeomorphic transformation applied to the digit $4$. 
It is important to note that the linearity of a transformation of pixel coordinates of a given image does not imply a linear transformation of the given image. For instance, the translation of an image is a linear operator in the coordinate space but highly nonlinear in the high-dimensional image space \cite{rao1999learning}.

\subsection{A Parametric Approach to Diffeomorphisms}

We consider as diffeomorphism a (smooth) mapping from $\mathbb{R}^2$ to $\mathbb{R}^2$ that transforms the coordinates of an input (image) to produce a deformed one \cite{younes2010shapes}. We choose to parametrize such a mapping by using a polyharmonic spline, specified by finitely many parameters. 
In particular, we consider the Thin-Plate-Spline (TPS) interpolation technique \cite{duchon1976interpolation}, which is also one of the most widely used transformations approximation methods in image registration problems dealing with nonlinear geometric differences \cite{nejati2010fast,bookstein1989principal}. 
%In particular, we consider the Thin-Plate-Spline (TPS) interpolation technique \cite{duchon1976interpolation}. The TPS is the radial basis function that minimizes an interpolation problem subject to the integral bending energy. 
% 
Essential to our work, the TPS is the biharmonic radial basis function that produces smooth surfaces from $\mathbb{R}^2$ to $\mathbb{R}^2$, which are infinitely differentiable \cite{morse2005interpolating}. In particular, it provides a transformation map that minimizes the volume described by the sum of the squares of the second derivative terms of the interpolant function, also referred to as the Biharmonic equations without boundary conditions \cite{sastry2015thin}. In particular,
We refer the reader to App.~\ref{app:TPS} for details regarding this method.

%It produces smooth surfaces, which are infinitely differentiable, and is one of the most widely used transformations approximation methods in image registration problems dealing with nonlinear geometric differences \cite{nejati2010fast,bookstein1989principal}. In particular, the TPS provides the transformation that minimizes the volume described by the sum of the squares of the second derivative terms, which corresponds to the trace of the square of the Hessian matrix, also referred to as the Biharmonic equations without boundary conditions \cite{sastry2015thin}. The reader can refer to App.~\ref{app:TPS} for details regarding this method. 
%
%The reader can refer to App.~\ref{app:TPS} for details regarding this method. 
%
%To produce such an approximation, the TPS finds the transformation that requires the least amount of bending energy. In particular, the TPS provides the transformation that minimizes the volume described by the sum of the squares of the second derivative terms, which corresponds to the trace of the square of the Hessian matrix, also referred to as the Biharmonic equations without boundary conditions \cite{sastry2015thin}. The reader can refer to App.~\ref{app:TPS} for details regarding this method. 
In this work, we consider as learnable parameters of the TPS a set of $2$-dimensional coordinates, called landmarks, and denoted by $\nu$. Given a set of landmarks, the TPS provides the parameters of the radial basis function performing the transformation map. That is, the euclidean plane is bent according to the learned landmarks.
While we optimize the landmarks directly, an alternative approach proposed in \cite{jaderberg2015spatial,li2017dense} predicted them using a deep neural network. 
In Fig.~\ref{fig:orig_to_diffeo}, we can see on the bottom right, the grid associated with the $\ell=6^2$ landmarks that enable the diffeomorphic transformation of the $4$. 

The TPS transformation of an image $x \in \mathbb{R}^n$, with $\ell$-landmarks, is written as
\setlength{\abovedisplayskip}{4pt}
\setlength{\belowdisplayskip}{4pt}
\begin{equation}
    \text{TPS}_{\ell}(x;\nu)~~,
    \label{eq:tps_notation}
\end{equation}
with $\nu \in \mathbb{R}^{2\ell}$, that is, $\ell$, $2-$dimensional coordinates.

\section{Deformation Invariant $K$-means}
\label{sec:RAI-Kmeans}
We now introduce our main contribution: the DI $K$-means algorithm. 
% Specifically, we introduce the DI $K$-means algorithm. 
The DI $K$-means combines the $K$-means' centroids-based clustering approach with a measure of similarity that we design to be invariant to diffeomorphisms. 
%diffeomorphic deformations

\subsection{Diffeomorphism-aware Similarity Measure}

Given a set of images $\{x_i \}_{i=1}^{N}$, with $x_i \in \mathbb{R}^{n}$, the $K$-means algorithm aims at grouping the data into $K$ distinct clusters defining the partition $\mathcal{C} = \left \{ C_k \right \}_{k=1}^{K}$, with $\cup_k C_k=\{x_i \}_{i=1}^{N}$ and $C_i\cap C_j = \emptyset, \forall i\neq j$. Each cluster $C_k$ of the partition is represented by a centroid $\mu_k \in \mathbb{R}^n, \forall k \in \{1,\dots, K\}$.

The goal of the DI $K$-means is to find the centroids minimizing the following distortion error
\begin{equation}
\begin{aligned}
& \min_{\mathcal{C}, \mu_{1},\ldots, \mu_K, \forall k, \left \| \mu_k \right \|_2=1}
& & \sum_{k=1}^{K} \sum_{i: x_i \in C_k}   d(x_i, \mu_k)~~.
\label{eq:RAI_Kmeans}
\end{aligned}
\end{equation}
The assignment of an image $x_i$ to a cluster $C_k$ is achieved through the evaluation of the similarity measure, $d$, between the image and each centroid. An image $x_i$ belongs to cluster $C_l$ if and only if $l = \argmin_k d(x_i, \mu_k)$. While the standard $K$-means algorithm makes use of the Euclidean distance, i.e., $d(x_i, \mu_k)= \|x_i - \mu_k \|_2^2$, we instead propose to use the following deformation invariant similarity measure
\begin{align}
    &d(x_i,\mu_k) := \min_{\nu \in \mathbb{R}^{2\ell}} \left \| \text{TPS}_{\ell}(x_i;\nu) - \mu_k \right \|_2^2~~.  \label{eq:def_invariant}  
\end{align}
This similarity measure characterizes the fitting of the TPS transformation of the image $x_i$ to best fit the centroid $\mu_k$ in the least-square sense. Note that it is a \textit{Quasipseudosemimetric}, see App.~\ref{app:proofs} for details and proof.
This measure requires solving a non-convex optimization problem. It can be achieved in practice by exploiting the differentiability of the TPS with respect to the landmarks $\nu$. As a result, we can learn the transformation by performing gradient-descent based optimization \cite{kingma2014adam}; further details regarding this optimization are given in Appendix~\ref{app:impl_details} as well as solutions to facilitate optimization of the non-convex objective by exploiting the image manifold geometry.

The essential property of the measure we propose is its invariance to deformations, formal proofs and definitions  in Appendix~\ref{app:invariance}. 

\subsection{Learning the DI $K$-means}

Solving the optimization problem in Eq.~\ref{eq:RAI_Kmeans}, similarly to $K$-means, is an NP-hard problem. A popular tractable solution nonetheless exists and is known as the two-step Lloyd algorithm \cite{lloyd1982least}. 

In the DI $K$-means, the first step of the Lloyd algorithm consists of assigning the images to a cluster using the newly defined measure of similarity in Eq.~\ref{eq:def_invariant}~. The second step, is the update of the centroids using the previously determined cluster assignment. It corresponds to the result of the optimization problem: $\argmin_{\mu_k: \left \| \mu_k \right \|=1} \sum_{i: x_i \in C_k} d(x_i,\mu_k)$, which solution is the following Proposition~\ref{prop2}.
\begin{prop}
\label{prop2}
The centroids update of the DI $K$-means algorithm are given by
\begin{equation}
\normalfont   \mu_k^{\star} \propto \frac{1}{\left | C_k \right |}  \sum_{i: x_i \in C_k} \text{TPS}_{\ell}(x_i;\nu_{i,k}^{\star}),  \: \: \forall k 
   \label{eq:update}
\end{equation}
where $\left | C_k \right |$ denotes the cardinal of the set $C_k$, $\normalfont \nu_{i,k}^{\star}$ is the set of parameters of the TPS that best transforms the image $x_i$ to the centroid $\mu_k$ defined in Eq.~\ref{eq:def_invariant}, and $\propto$ is defines the proportionality relation (proof in App.~\ref{proof:prop2}).%per the optimization problem
\end{prop}

The averaging in Eq.~\ref{eq:update} is performed on the transformed version of the images. DI $K$-means thus considers the topology of the images' ambient space. A pseudo-code of the centroid update Eq.~\ref{eq:update} is presented in Algo.~\ref{algo:update}. 
 \begin{algorithm}
 \caption{Centroids Updates of DI $K$-means}
 \begin{algorithmic}[1]
 \renewcommand{\algorithmicrequire}{\textbf{Input:} }
 \renewcommand{\algorithmicensure}{\textbf{Output:} }
 \REQUIRE  Cluster  $C_k$,  TPS parameters $\left \{ \nu_{i,k}^{\star} \right \}_{i: x_i \in C_k}$% , k \in \left \{1,\dots,K \right \}
 \ENSURE  Centroids update $ \mu_{k}^{\star}$
 \STATE Initialize $\mu_k = 0$
 \FOR{$i: x_i \in C_k$}  
  \STATE Compute $\mu_k = \mu_k + \text{TPS}_{\ell}(x_i;\nu_{i,k}^{\star})$, \; Eq.~\ref{eq:update}
 \ENDFOR
 \STATE $\mu_k^{\star} = \frac{\mu_k}{| C_k |}$
 \end{algorithmic} 
 \label{algo:update}
 \end{algorithm}

The training of the DI $K$-means, which aims at minimization the distortion error Eq.~\ref{eq:RAI_Kmeans} is done by alternating between the two steps detailed above, until a stopping criterion is met, and is summarized in Algo.~\ref{algo:algo}. 
%
% , following the two-step approach described above, we can evaluate and optimize the distortion error Eq.~\ref{eq:RAI_Kmeans}. The centroids are updated according to the algorithm described in Algo.~\ref{algo:update}. The learning of the transformations parameters and the update of the centroids are then performed iteratively and
%
%
%
 \begin{algorithm}
 \caption{Deformation Invariant $K$-means}
 \begin{algorithmic}[1]
 \renewcommand{\algorithmicrequire}{\textbf{Input:}  }
 \renewcommand{\algorithmicensure}{\textbf{Output:} }
 \REQUIRE Initial centroids $\mu_k$, dataset $\left \{x_i \right \}_{i=1}^N$
 \ENSURE  Cluster partition  $\left \{C_k \right \}_{k=1}^{K}$
 \REPEAT
 \FOR{$i=1$ to $N$}  
  \FOR{$k=1$ to $K$}
  \STATE Compute and store $d(x_i,\mu_k)$ by solving Eq.~\ref{eq:def_invariant}
  \ENDFOR
  \STATE Assign $x_i$ to $C_l$ where $l = \argmin_{k} d(x_i,\mu_k)$
  \ENDFOR
 \STATE Update the centroid $\mu_k$ using Algo.~\ref{algo:update}
 \UNTIL{Stopping criterion}
 \end{algorithmic} 
 \label{algo:algo}
 \end{algorithm}

The update Eq.~\ref{eq:update} induced by our similarity measure alleviates a fundamental limitation of the standard $K$-means when applied in the pixel space of the images.
In fact, in the standard $K$-means, the average of the data belonging to a cluster $C_k$, $\frac{1}{ | C_k|}\sum_{i: x_i \in  C_k} x_i$, consists of an averaging in the pixel space, which as a result does not account for the non-Euclidean geometry of the image manifold \cite{klassen2004analysis,srivastava2005statistical}.  For example, when considering faces, the naive ``average face" of an individual over different poses based on the Euclidean distance does not correspond to a recognizable picture of that individual. 

\vspace{-.3cm}
\subsection{Computational Complexity \& Parameters}
The time complexity of DI $K$-means is $O(NK ( \ell^3 + \ell n))$. In fact, the DI $K$-means computes a TPS of computational complexity $O(\ell^3 + \ell n)$ for each sample of the $N$ samples and each of the $K$ centroids, as in Eq.~\ref{eq:def_invariant}.
The number of parameters of the model is $ 2\ell  \times N \times K$, it depends on the number of samples, clusters, and landmarks.  

\begin{figure}[!h]
\begin{minipage}{.02\linewidth}
\rotatebox{90}{Time (min.)}
\end{minipage}
\begin{minipage}{.47\columnwidth}
    \centering
    \includegraphics[width=\linewidth]{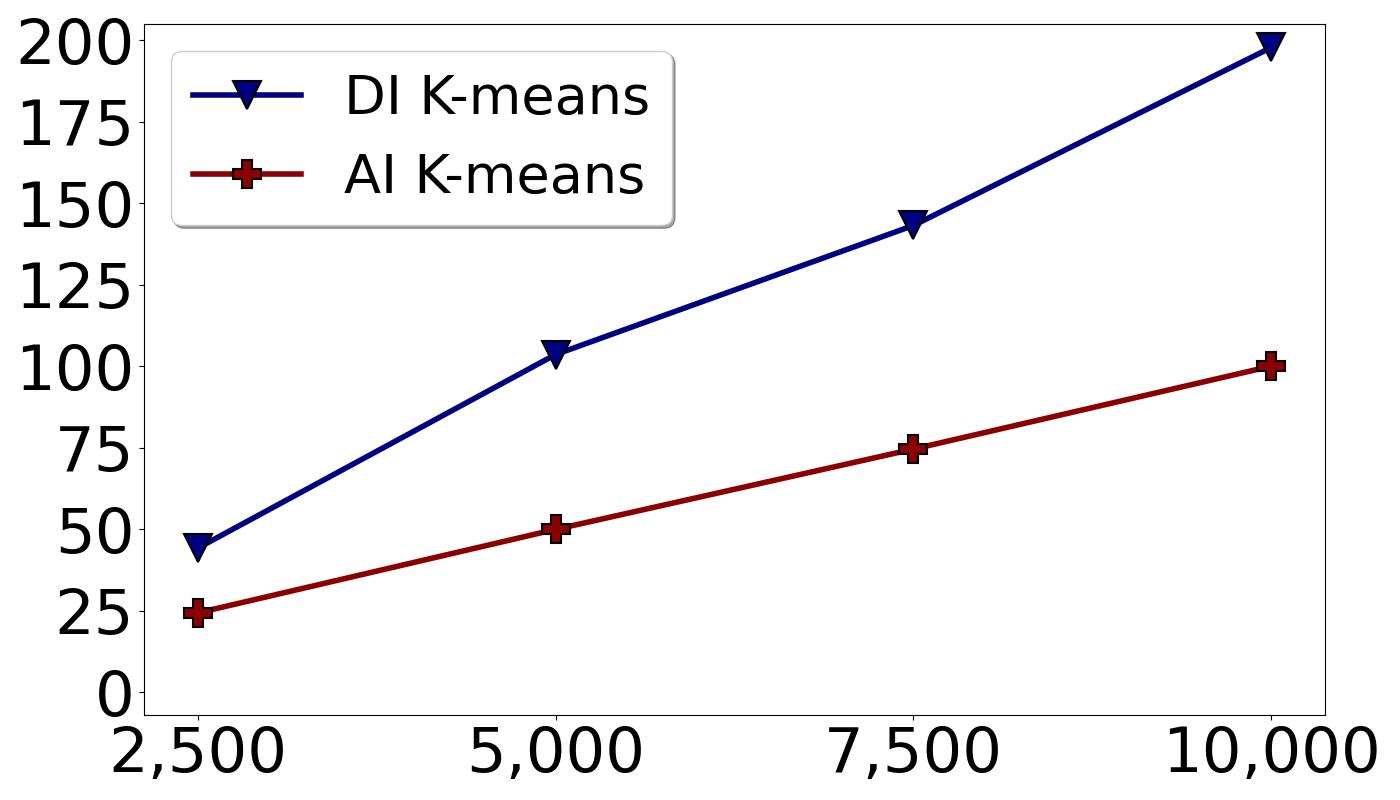}
\end{minipage}
\begin{minipage}{.47\columnwidth}
    \centering
    \includegraphics[width=\linewidth]{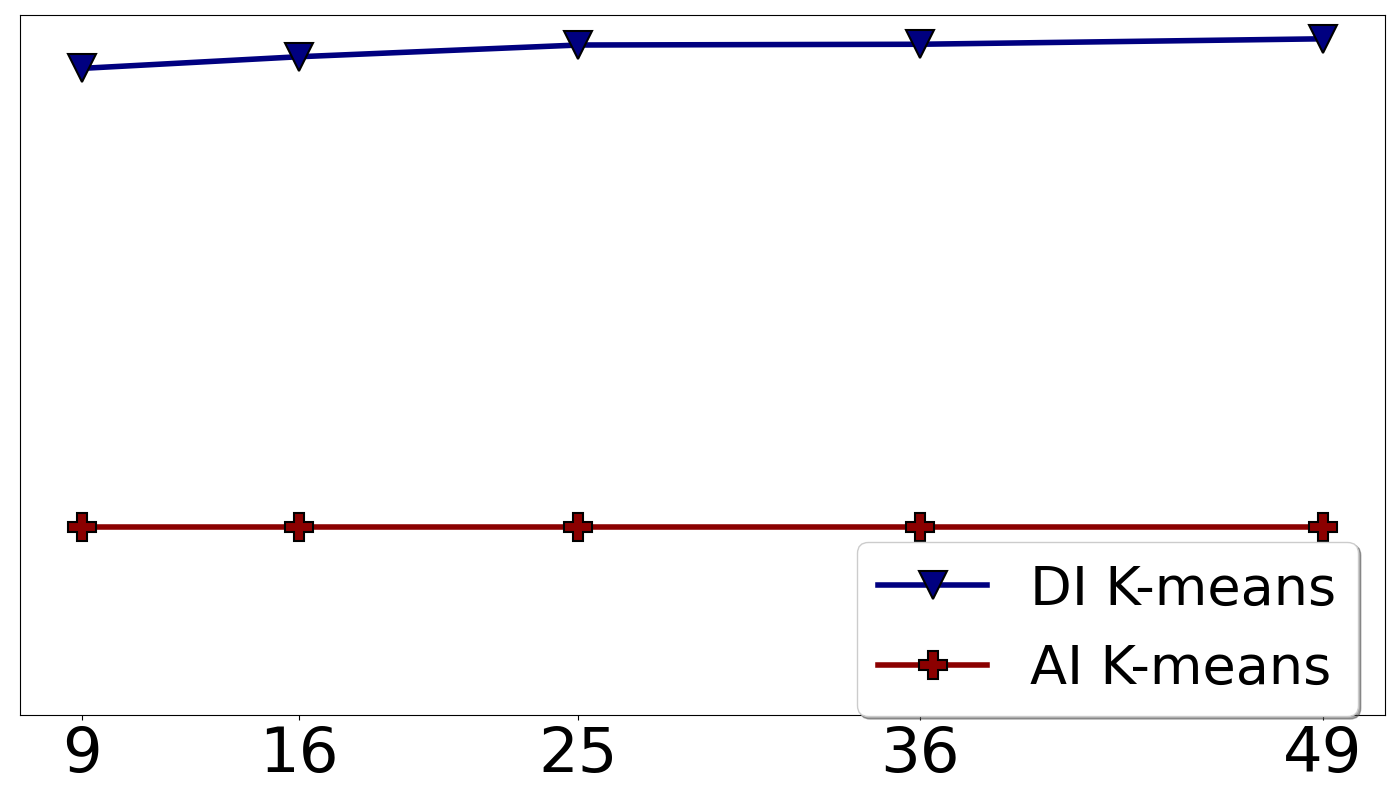}
\end{minipage}
\begin{minipage}{\columnwidth}
\hspace{1.1cm} Training Size ($N$) \hspace{.9cm} Number Landmarks ($\ell$)
\end{minipage}
\caption{ \textbf{Computational Training Time} - Comparison between our DI $K$-means and the AI $K$-means computational times on the Arabic Characters dataset. The input pixel size is $n=1024$. (\textit{Left}) shows the computational time for varying training set size and $\ell=7^2$. (\textit{Right}) shows the variations as a function of the number of landmarks, $\ell$, for $N=10,000$. Since the AI $K$-means does not use the TPS algorithm, its computational time is constant as a function of the number of landmarks.}
\label{fig:computational_time}
\end{figure}

%
%The time complexity of the TPS is $O(\ell^3 + \ell n)$, with $n$ the number of pixels in the image and $\ell$ the number of landmarks. DI $K$-means computes a TPS for each sample and each centroid (recall Eq.~\ref{eq:def_invariant}) leading to its time complexity being $O(NK ( \ell^3 + \ell n))$. 
%
To speed up the computation, we 1) pre-compute the matrix inverse responsible for the dominating cubic term, see Appendix~\ref{app:TPS} for implementation details regarding the TPS, and 2) implement DI $K$-means on GPU with SymJAX \cite{balestriero2020symjax} where high parallelization renders the practical computation time near constant with respect to the number of landmarks as we depict in Fig.~\ref{fig:computational_time}.

\begin{figure*}[p]
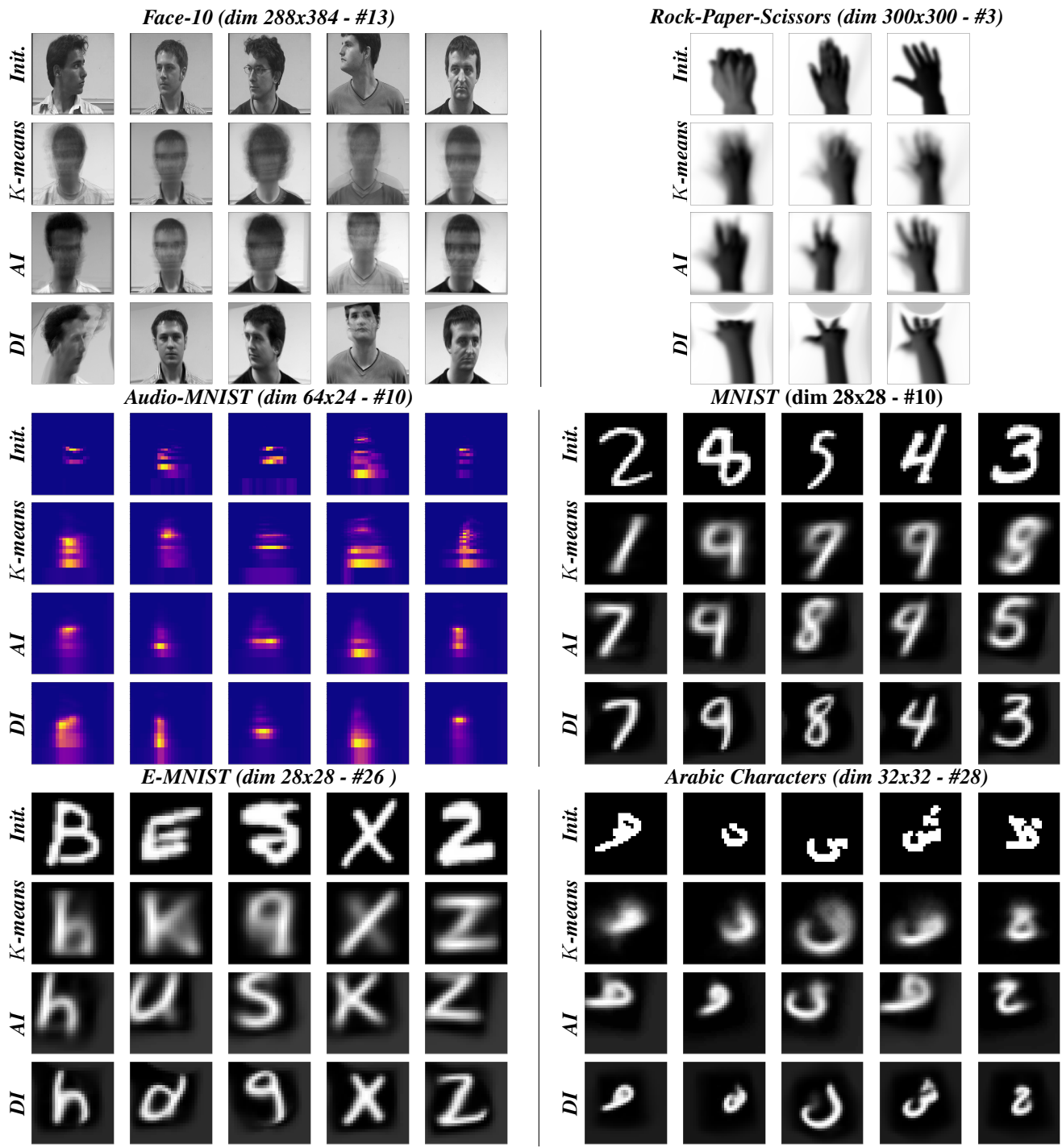


\begin{minipage}{.49\linewidth}
    \centering
    \textbf{\textit{Face-10 (dim 288x384 - \#13)}}
\end{minipage}
\begin{minipage}{.49\linewidth}
        \centering
        \hspace{1cm}
    \textbf{\textit{Rock-Paper-Scissors (dim 300x300 - \#3)}}
\end{minipage}

\begin{minipage}{.013\linewidth}
\rotatebox{90}{ \hspace{.1cm} \textbf{\textit{DI \hspace{.75cm} AI \hspace{.65cm} $K$-means \hspace{.44cm} Init.}}}
\end{minipage}
\begin{minipage}{0.46\linewidth}
\foreach \c in {0,1,3,4,7}{
    \begin{minipage}{0.18\linewidth}
    \includegraphics[width=\linewidth]{images/best_facepos_centroid_init\c.png}\\
    \includegraphics[width=\linewidth]{images/kmean_facepos_centroid_final\c.png}\\
    \includegraphics[width=\linewidth]{images/aff_facepos_centroid_final\c.png}\\
    \includegraphics[width=\linewidth]{images/best_facepos_centroid_final\c.png}
    \end{minipage}
}
\end{minipage}
\hfill \hspace{-.35cm}\vline\hfill
\begin{minipage}{0.46\linewidth}
\centering
\begin{minipage}{.008\linewidth}
\rotatebox{90}{ \hspace{.1cm} \textbf{\textit{DI \hspace{.75cm} AI \hspace{.65cm} $K$-means \hspace{.44cm} Init.}}}
\end{minipage}
\foreach \c in {0,...,2}{
    \begin{minipage}{0.18\linewidth}
    \includegraphics[width=\linewidth]{images/best_rockpaper_centroid_init\c.png}\\
    \includegraphics[width=\linewidth]{images/kmean_rockpaper_centroid_final\c.png}\\
    \includegraphics[width=\linewidth]{images/aff_rockpaper_centroid_final\c.png}\\
    \includegraphics[width=\linewidth]{images/best_rockpaper_centroid_final\c.png}
    \end{minipage}
}
\end{minipage}

\begin{minipage}{.49\linewidth}
    \centering
    \textbf{\textit{Audio-MNIST (dim 64x24 - \#10)}}
\end{minipage}
\begin{minipage}{.49\linewidth}
        \centering
        \hspace{1cm}
    \textbf{\textit{MNIST} (dim 28x28 - \#10)}
\end{minipage}

\begin{minipage}{.013\linewidth}
\rotatebox{90}{ \hspace{.1cm} \textbf{\textit{DI \hspace{.75cm} AI \hspace{.65cm} $K$-means \hspace{.44cm} Init.}}}
\end{minipage}
\begin{minipage}{0.46\linewidth}
\foreach \c in {0,...,4}{
    \begin{minipage}{0.18\linewidth}
    \includegraphics[width=\linewidth]{images/best_audiomnist_centroid_init\c.png}\\
    \includegraphics[width=\linewidth]{images/kmean_audiomnist_centroid_final\c.png}\\
    \includegraphics[width=\linewidth]{images/aff_audiomnist_centroid_final\c.png}\\
    \includegraphics[width=\linewidth]{images/best_audiomnist_centroid_final\c.png}
    \end{minipage}
}
\end{minipage}
\hfill \hspace{-.1cm}\vline\hfill
\begin{minipage}{.013\linewidth}
\rotatebox{90}{ \hspace{.1cm} \textbf{\textit{DI \hspace{.75cm} AI \hspace{.65cm} $K$-means \hspace{.44cm} Init.}}}
\end{minipage}
\begin{minipage}{0.46\linewidth}
\foreach \c in {0,2,6,7,9}{
    \begin{minipage}{0.18\linewidth}
    \includegraphics[width=\linewidth]{images/best_mnist_centroid_init\c.png}\\
    \includegraphics[width=\linewidth]{images/kmean_mnist_centroid_final\c.png}\\
    \includegraphics[width=\linewidth]{images/aff_mnist_centroid_final\c.png}\\
    \includegraphics[width=\linewidth]{images/best_mnist_centroid_final\c.png}
    \end{minipage}
}
\end{minipage}

\begin{minipage}{.49\linewidth}
    \centering
    \textbf{\textit{E-MNIST (dim 28x28 - \#26 )}}
\end{minipage}
\begin{minipage}{.49\linewidth}
        \centering
        \hspace{1cm}
    \textbf{\textit{Arabic Characters (dim 32x32 - \#28)}}
\end{minipage}

\begin{minipage}{.013\linewidth}
\rotatebox{90}{ \hspace{.1cm} \textbf{\textit{DI \hspace{.75cm} AI \hspace{.65cm} $K$-means \hspace{.44cm} Init.}}}
\end{minipage}
\begin{minipage}{0.46\linewidth}
\foreach \c in {0,2,3,4,7}{
    \begin{minipage}{0.18\linewidth}
    \includegraphics[width=\linewidth]{images/best_emnist_centroid_init\c.png}\\
    \includegraphics[width=\linewidth]{images/kmean_emnist_centroid_final\c.png}\\
    \includegraphics[width=\linewidth]{images/aff_emnist_centroid_final\c.png}\\
    \includegraphics[width=\linewidth]{images/best_emnist_centroid_final\c.png}
    \end{minipage}
}
\end{minipage}
\hfill \hspace{-.1cm}\vline\hfill
\begin{minipage}{.013\linewidth}
\rotatebox{90}{ \hspace{.1cm} \textbf{\textit{DI \hspace{.75cm} AI \hspace{.65cm} $K$-means \hspace{.44cm} Init.}}}
\end{minipage}
\begin{minipage}{0.46\linewidth}
\foreach \c in {0,2,3,4,9}{
    \begin{minipage}{0.18\linewidth}
    \includegraphics[width=\linewidth]{images/best_arabchar_centroid_init\c.png}\\
    \includegraphics[width=\linewidth]{images/kmean_arabchar_centroid_final\c.png}\\
    \includegraphics[width=\linewidth]{images/aff_arabchar_centroid_final\c.png}\\
    \includegraphics[width=\linewidth]{images/best_arabchar_centroid_final\c.png}
    \end{minipage}
}
\end{minipage}
\caption{ \textbf{Centroids} - The name of the datasets considered is written on top of each subfigure, and \# corresponds to the number of classes. For each dataset, we depict the centroids at initialization in the top row. The centroids learned by $K$-means are shown in the \textit{\nth{2} row}, by the Affine invariant $K$-means in the \textit{\nth{3} row}, and by our DI $K$-means in the \textit{\nth{4} row}. 
By comparing the results of the AI $K$-means (\textit{\nth{3} row}) with the standard $K$-means (\textit{\nth{2} row}), we can see that using only affine transformations slightly improves the $K$-means centroids and reduces the superposition issue that $K$-means suffers from. 
%We can see how the ability of the method to adapt the data based solely on affine transformations  slightly improves the $K$-means centroids and reduces the superposition issue that $K$-means suffers from. 
By comparing the results of our DI $K$-means (\textit{\nth{4} row}) with the other methods, it is clear that using the more general class of transformation that are diffeomorphisms, via the TPS, significantly improves the centroids, making them sharper and removing the issue related to the non-additiveness of images. Note that $K$-means iteratively updates the centroids and cluster assignments, as such, the class associated to a specific centroid usually changes during training .
%By introducing the proposed more general class of transformation via the TPS, we see that the final centroids are much crisper as the data have been successfully mapped through their corresponding centroids prior to performing the centroid update, Eq.~\ref{eq:update}.
}
    \label{fig:centroids}
\end{figure*}

\begin{figure*}[ph!]
\begin{minipage}{.32\linewidth}
    \centering
\textbf{\textit{Raw Data}}
\end{minipage}
\begin{minipage}{.32\linewidth}
    \centering
\textbf{\textit{Affine Invariant}}
\end{minipage}
\begin{minipage}{.32\linewidth}
    \centering
\textbf{\textit{Deformation Invariant}}
\end{minipage}
\begin{minipage}{0.03\linewidth}
\rotatebox{90}{\hspace{.6cm} \textbf{MNIST \#10} \hspace{2.cm} \textbf{Audio-MNIST \#10} \hspace{1.1cm} \textbf{Face-10 \#13} \hspace{1.5cm} \textbf{Rock-Paper-Scissors \#3}}
\end{minipage}
\begin{minipage}{0.95\linewidth}
    \centering
    \includegraphics[width=\linewidth]{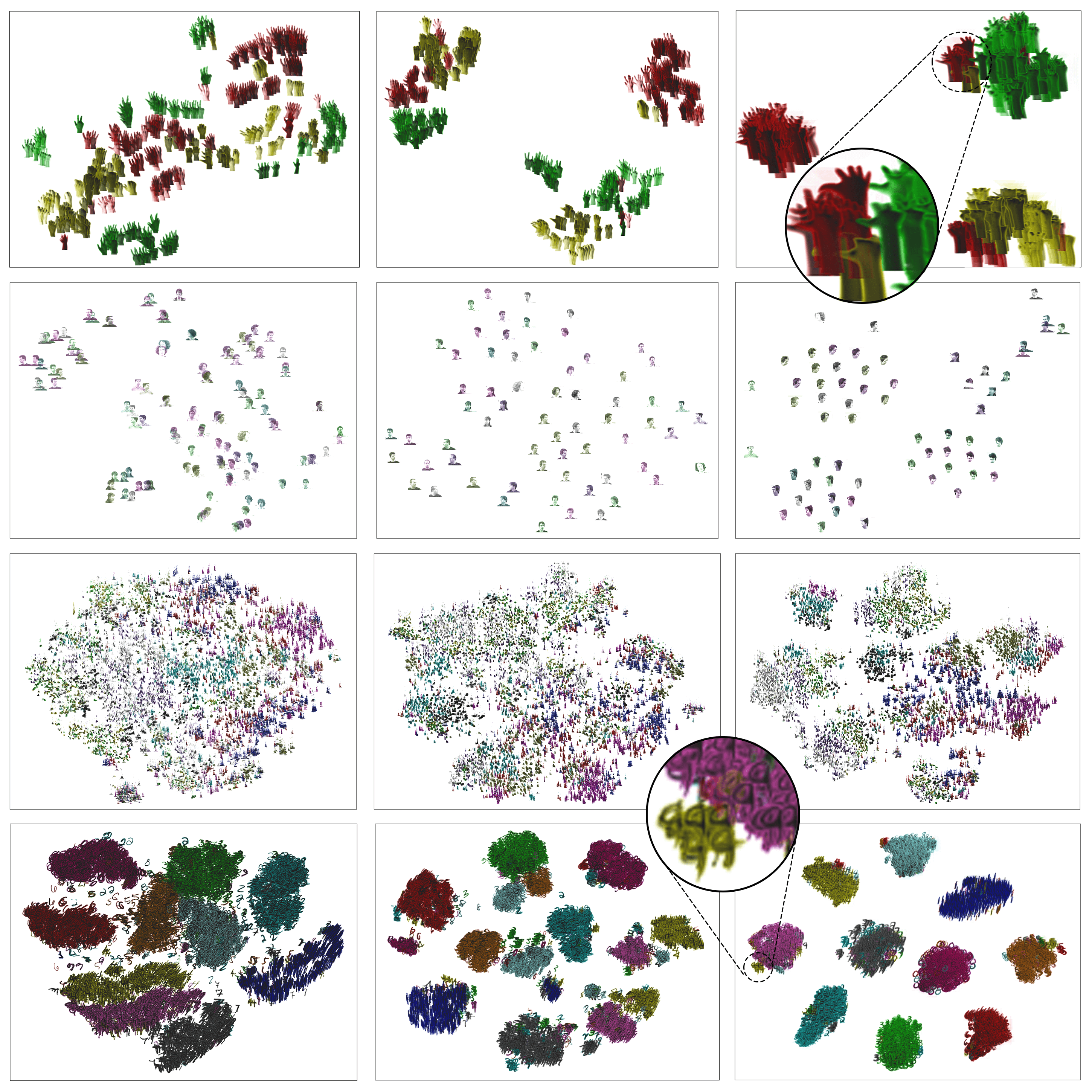}
\end{minipage}
    \caption{\textbf{$2$-dimensional t-SNE} - (\# denotes the number of clusters) - We suggest the reader to zoom in the plots to best appreciate the visualization. - The raw data (\textit{left column}), the affinely transformed data using the AI distance, i.e., we extract the best affine transformation of the data that corresponds to the centroid it was assigned and perform the t-SNE on these affinely transformed data, (\textit{middle column}), the data transformed with respect to the TPS as per Eq.~\ref{eq:def_invariant}, i.e., the same process as previously mentioned but we consider the transformation induced by the TPS, and then perform the dimension reduction on these transformed data, (\textit{right column}). Each row corresponds to a different datasets, Rock-Paper-Scissor, Face-10, AudioMNIST, and MNIST  are depicted from the top to bottom row. For all the figures, the colors of the data represent their ground truth labels. We observe that across datasets, both the affine transformation learned on the data and the TPS transformation help to define more localized clusters. One can observe that for the Face-10 dataset, while the dataset contains $13$ clusters, we can see that the DI $K$-means induced transformations lead to a $2$-dimensional space where the faces are clustered $3$ majors orientations. The top left cluster corresponds to faces pointing left, the bottom one face pointing right, and the bottom right one face pointing front.}
\label{fig:tsne}
\end{figure*}

\vspace{-.3cm}
\section{Experimental Setup and Cross Validation}
\label{sec:experiments}

% We evaluate and compare the DI $K$-means with competing unsupervised models on various datasets described in Appendix~\ref{ap:data}. 
%
In this section, we detail the experimental setting followed to evaluate the performance of our model as well as important details on the cross-validation of the various models. 
It is important to note that our model operates in the unsupervised settings, thus we do not compare our method with models trained leveraging the cluster membership labels. Also, for all clustering algorithms, the number of clusters is set to be the number of classes the dataset contains. 
The various datasets used are described in Appendix~\ref{ap:data}. 
%
% For each dataset, we compare our method with the $K$-means algorithm, the Affine Invariant $K$-means \cite{fitzgibbon2002affine}, which only considered affine transformations. 
%
% We also compare DI $K$-means with various clustering techniques using Deep Learning.

\subsection{Evaluation Metric}

For all the experiments, the accuracy is calculated using the metric proposed in \cite{yang2010image} and defined as
\setlength{\abovedisplayskip}{3pt}
\setlength{\belowdisplayskip}{3pt}
%\small
\begin{align}
 \text{Accuracy} =    \max_{m} \tfrac{1}{N} \sum_{i=1}^{N} 1_{ \left \{ l_i = m(\hat{l}_i) \right \}}~~,
\label{eq:cluster} 
\end{align}%\normalsize
where $l_i$ is the ground-truth label, $\hat{l}_i$ the cluster assignment and $m$ all the possible one-to-one mappings between clusters and labels. 
The results in Table~\ref{table:compare} are taken as the best score on the test set based on the ground truth labels among $10$ runs as in \cite{xie2016unsupervised}.

\subsection{Competing Models}
\label{sec:compete}
We compare our model with well-known clustering techniques using deep neural networks. We performed experiments for the VaDE \cite{jiang2016variational} and DEC \cite{xie2016unsupervised} using the code made publicly available by the authors. 
%state-of-the-art
We use the annotation (MLP) as a reference to the MLP architecture used in their experiments (see App.~\ref{app:arch} for details). 
To fairly compare our model to the DEC and VaDE models, we proposed a convolutional architecture to the DEC and VaDE networks, denoted by DEC (Conv) and VaDE (Conv) (see App.~\ref{app:arch} for details). 
Finally, we evaluate the performance of an augmented $K$-means algorithm trained using the features extracted by an Autoencoder, denoted by AE $+$ $K$-means in the following. 
%(Conv)

The parameters of the different models mentioned above are learned by stochastic gradient descent (Adam optimizer \cite{kingma2014adam}). In all the experiments, the learning rate are cross-validated according to $[10^{-4}, 5\times10^{-4}, 10^{-3}, 5\times 10^{-3}, 10^{-2},5\times 10^{-2} ]$. The internal parameters that are model dependant, e.g., the number of pre-training epoch and the update intervals, are also cross-validated.  
% As in our method, each model is initialized $10$ times and the best result is reported. 

We also compare our DI $K$-means to the closely related $K$-means and affine invariant $K$-means, denoted by AI $K$-means in the following. The AI $K$-means was designed to include only affine transformations. 
For each run, all three $K$-means algorithms start from the same initial centroids using the $K$-means$++$ algorithm developed by \cite{arthur2006k} to speed up the convergence of the $K$-means algorithm. 

\subsection{Cross Validation Settings}

\begin{figure}[!h]
\begin{minipage}{.04\columnwidth}
\rotatebox{90}{\hspace{.45cm} Accuracy}
\end{minipage}
\begin{minipage}{.96\columnwidth}
    \centering
    \includegraphics[trim=0 0.5cm 0 0,width=1\columnwidth]{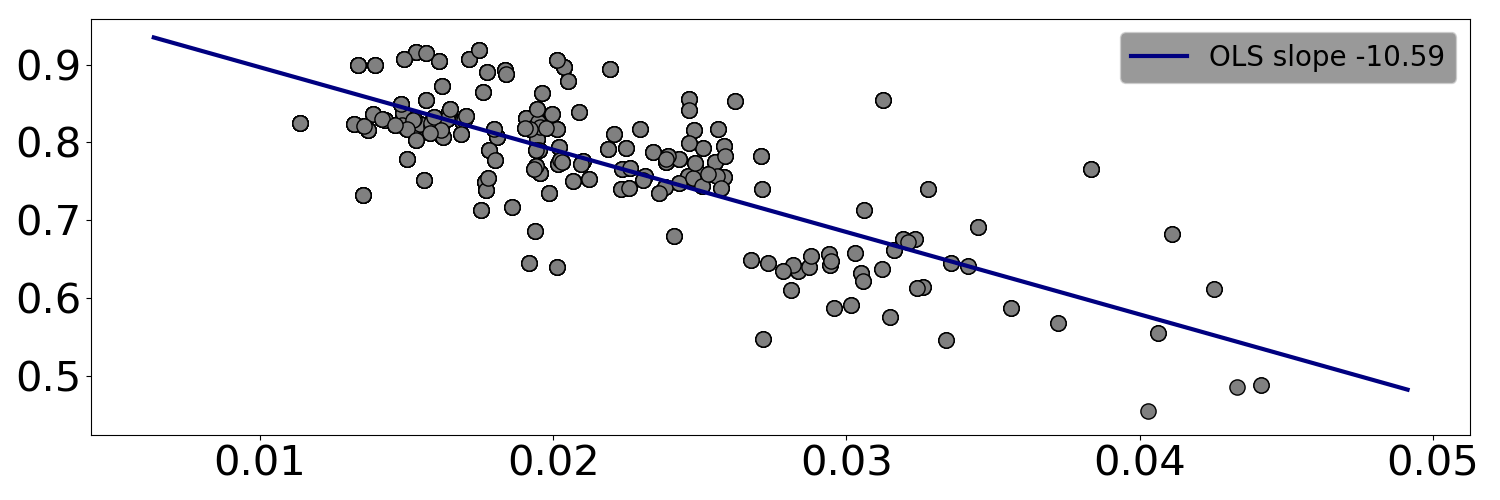}
\begin{minipage}{\columnwidth}
\centering
\hspace{.5cm} Distortion Error
\end{minipage}
\end{minipage}
\caption{\textbf{Accuracy vs Distortion Error} -  
Clustering accuracy, Eq.~\ref{eq:cluster}, of DI $K$-means algorithm on the MNIST dataset as a function of the distortion error, Eq.~\ref{eq:RAI_Kmeans}, using the similarity measure, Eq.~\ref{eq:def_invariant}. Each gray dot is associated with a specific set of hyper-parameters, e.g., learning rate, number of landmarks for the TPS. 
%, and random initialization 
%The x-axis denotes the distortion error, Eq.~\ref{eq:RAI_Kmeans}, using the similarity measure, Eq.~\ref{eq:def_invariant}, and the y-axis the clustering accuracy, defined in Eq.~\ref{eq:cluster}, for the DI $K$-means algorithm applied on the MNIST dataset. Each gray dot corresponds to specific hyper-parameters: learning rate, number of landmarks for the TPS, and random initialization. 
%
The accuracy is negatively correlated to the distortion error (see the blue line corresponding to the ordinary least square fit), indicating that the distortion error is an appropriate metric to cross-validate the hyper-parameters of the DI $K$-means algorithm. }
\label{fig:accu_vs_dist}
\end{figure}

Our model requires the cross-validation of hyper-parameters related to the learnability of the transformation within the similarity measure, Eq.~\ref{eq:def_invariant}. 
However, the clustering framework does not allow the use of label information to perform the cross-validation of the parameters. We thus need to find a proxy for it to determine the optimal model parameters. 
Interestingly, the distortion error related used in the DI $K$-means, Eq.~\ref{eq:RAI_Kmeans}, appears to be negatively correlated to the accuracy, as displayed in Fig.~\ref{fig:accu_vs_dist}. 
The use of the distortion error is commonly used as a fitness measure in $K$-means for example when cross-validating the number of clusters. 
% to employ

%
% As we stand in the framework of clustering, the cross-validation can not be performed using the existing label. Therefore, the accuracy can not be leveraged as to select the appropriate model. Interestingly, the distortion error related to the DI $K$-means, defined in Eq.~\ref{eq:RAI_Kmeans}, is negatively correlated to the accuracy, see Fig.~\ref{fig:accu_vs_dist}. Therefore, the cross-validation of the hyper-parameters of the model can be achieved using the distortion error directly, which does not require any label.

We cross-validate the number of landmarks, $\ell$, which defines the resolution of the transformation, which we optimize over the following grid, $\left [3^2,4^2,5^2,6^2,7^2,8^2 \right ]$. 
Then, the learning of the landmarks, $\nu$, is done via Adam optimizer.
The learning rate is picked according to $\left [10^{-4},5\times 10^{-4},10^{-3},5\times 10^{-3},10^{-2},5\times 10^{-2} \right ]$. 
We train our method for $150$ epochs for all the datasets, with batches of size $64$. 
As for $K$-means and AI $K$-means, the centroids' initialization of the DI $K$-means is performed by the $K$-means$++$ algorithm. 
Importantly, the same procedure is applied to all datasets. 
% $\left [0.0001,0.0005,0.001,0.005,0.01,0.05 \right ]$.
% developed by \cite{arthur2006k} to speed up the convergence of the $K$-means algorithm. 

Note that during the training, both the similarity measure in Eq.~\ref{eq:def_invariant} and the clustering update are performed, Eq.~\ref{eq:cluster}. During the algorithm's testing phase, the centroids remain fixed, and only the similarity measure is trained to assign each testing datum to a cluster. For a given testing sample, we find its best fit with respect to all centroids $\mu_k^{\star}, \forall k$, and assign it to the closest according to the DI similarity measure, Eq.~\ref{eq:def_invariant}.

\section{Results and Interpretation}
In this section, we report and interpret the results obtained by our DI $K$-means and competing models.

\begin{table*}[t!]
\centering
\caption{Clustering Results in $\%$ of the test set accuracy Eq.~\ref{eq:cluster} - the number of clusters is denoted by \# next to the dataset name.}
 %\begin{adjustbox}{width=\textwidth}
 \setlength\tabcolsep{1.25pt}
\begin{tabular}{|l|c|c|c|c|c|c|c|c|c|c|}
\hline
\hspace{.8cm} \textit{\textbf{Models}} & \textit{\textbf{Deep}} & \textit{\textbf{Affine}}    &  \textit{\textbf{Diffeo.}} & \textit{\textbf{MNIST}} & \textit{\textbf{Audio}} & \textit{\textbf{E-MNIST}} & \textit{\textbf{Rock-Paper}} & \textit{\textbf{Face-10}} & \textit{\textbf{Arabic}} \\
 & \textit{\textbf{Learning}} & \textit{\textbf{MNIST \#10}}    &  \textit{\textbf{MNIST \#10}} & \textit{\textbf{\#10}} & \textit{\textbf{ MNIST \#10}} & \textit{\textbf{\#26}} & \textit{\textbf{-Scissors \#3}} & \textit{\textbf{\#13}} & \textit{\textbf{Char. \#28}} \\\hline
$K$-means (\cite{xie2016unsupervised})& \xmark  &   68.0 & 61.8 & 53.5 & 10.8 & 39.5 & 40.6 & 20.5 & 19.4   \\\hline
AI $K$-means & \xmark &  \textbf{100.0}  &  91.7 & 75.3 & 29.8 & 48.0& 72.8  & 31.2 & 30.25 \\ \hline
DI $K$-means (\textbf{Our}) &  \xmark  &  \textbf{100.0} & \textbf{99.2} & 92.5 & \textbf{41.6} & \textbf{65.3}  &  \textbf{86.6} &  \textbf{45.3} & \textbf{51.1} \\ \hline \hline
AE + $K$-means   &   \cmark  & 72.6  & 60.5 & 66.3 & 13.8 & 41.8 & 48.1 & 37.7 & 23.5 \\  \hline
DEC (MLP) (\cite{xie2016unsupervised})  &  \cmark &  84.3  & 77.6 & 84.3 & 10.9 & 55.4 & 46.2 & 33.6 & 24.2 \\\hline
DEC (Conv)  &  \cmark &  70.4  & 68.2 & 78.8  & 15.3 & 60.1  & 54.1 & 38.3 &  29.1 \\\hline
VaDE (MLP) (\cite{jiang2016variational}) &  \cmark &  68.8  & 65.4 & \textbf{94.1} & 11.3 & 20.3 & 50.3 & 36.4 & 26.2 \\\hline
VaDE (Conv) &  \cmark &  65.4  & 59.3 & 81.7 & 14.3 &  58.3 & 55.4  & 40.9 & 46.4 \\\hline
\end{tabular}
\label{table:compare}
%\end{adjustbox}
\end{table*}

\subsection{Clustering Accuracy}

We report in Table~\ref{table:compare} the accuracy of the various models considered on the different datasets. 
Our approach shows to outperform existing models on most datasets. Our model equals the performance of AI $K$-means on Affine MNIST and is only outperformed by VaDE (MLP) on MNIST. 
%
% For all the models, the number of clusters is set to the number of classes of the dataset. Note none of the model we compare with is using data augmentation. 
%
% Our approach shows to competes with the others methods and usually outperform them while not performing any embedding. 
%
Whereas the various deep learning approaches perform well on datasets for which their architectures were developed, e.g., MNIST and its derivatives: E-MNIST, Arabic Characters, they show limited performance on higher resolution datasets with a small number of samples, such as Rock-Paper-Scissors, Face-10 as well as the two toy examples (composed of only $700$ training data and $300$ testing data). 
%Note that deep learning approaches are performing well on datasets for which their architecture has been developed, in the present case, MNIST and its derivatives: E-MNIST, Arabic Characters, while performing poorly on higher resolution datasets with a small number of samples, such as Rock-Paper-Scissors, Face-10 as well as the two toy examples (composed of only $700$ training data and $300$ testing data). 

\subsection{Interpretability: Centroids Visualization }

We propose in Fig.~\ref{fig:centroids} to visualize the centroids obtained via $K$-means, AI $K$-means, and our DI $K$-means. Supplementary vizualisations are provided in Appendix~\ref{app:centroids}.
For each dataset, the first row shows the clusters after initialization from $K$-means$++$, as mentioned in Sec.~\ref{sec:compete} .
The three following rows show the centroids obtained via the $K$-means, AI $K$-means, and DI $K$-means algorithms, respectively. 

We observe that, for all datasets, the $K$-means centroids are not possible samples from the data. In that sense, the $K$-means algorithm does not consider the non-linear manifold structure of the data. In fact, its update rule consists in the average of the data belonging to each cluster in the pixel space, thus producing blurred and mixed centroids. 
%
%Across datasets, we observe that the Euclidean distance's update does not consider the manifold structure of the data as it corresponds to the average of the data belonging to each cluster. 
%
The AI $K$-means algorithm drastically reduces the blurriness of the centroids induced by such an averaging as it considers the average of affinely transformed data. 
However, our DI $K$-means produces the crispest centroids and does not introduce any ambiguity in between the different clusters. 
In fact, the update of our method, Eq.~\ref{eq:update}, takes into account the non-linear structure of the image manifold by taking the average over diffeomorphically transformed data. 

Interestingly, Fig.~\ref{fig:centroids} shows that even if at initialization multiple centroids assigned to the same class are attributed to different clusters, the DI $K$-means is able to recover this poor initialization thanks to its explicit manifold modeling and centroid averaging technique.
%the DI $K$-means modeling converge to a solution where each cluster corresponds to a different class, and is thus more suited for computer vision tasks. In fact, DI $K$-means proves able to discover centroids that best capture the variability in the data. 
%
%One can see in Fig.~\ref{fig:centroids} that even if the initalization is not correct, that is, multiple instances representing the same class are attributed to different clusters, the DI $K$-means corrects such mistake to discover centroids that best capture the variability in the data. 
%
For instance, in the Rock-Paper-Scissors dataset, although at initialization, two centroids correspond to the class paper, the DI $K$-means learns centroids of each of the three classes within this dataset. 
%
%For instance, in the case of the Rock-Paper-Scissors dataset, at the initialization, two centroids out of three are instances of the class paper, while the resulting DI $K$-means centroids are corresponding to the three classes within this dataset. 
%
In the Face-10 dataset, some centroids learned correspond to the rotation of the initialization, even in such extreme change of pose, the centroids remain crisp in most cases. 
\vspace{-.1cm}
\subsection{Interpretability: Embedding Visualization}

To get further insights into the diffeomorphism-awareness and disentangling capability of the DI $K$-means, we compare the $2$-dimensional projections of the data using t-SNE \cite{maaten2008visualizing}, of the $K$-means, AI $K$-means and DI $K$-means. Supplementary vizualisations are provided in Appendix~\ref{app:sup_tsne}.
The t-SNE visualizations, for both the AI and DI $K$-means, are obtained by extracting the optimal transformation that led to the assignment. 
%
%Precisely, for each image $x_i$, we compute $l = \argmin_k d(x_i, \mu_k)$  and extract the optimal parameter $\nu_{i,l}^{\star}$. 
%
%These transformation parameters are then used to obtain the transformed image that best fits its appropriate centroid, i.e., $ \text{TPS}_{\ell}(x_i;\nu_{i,l}^{\star})$. 
%
Then t-SNE is then applied to all the transformed data using the optimal transformation. 

We can observe in Fig.~\ref{fig:tsne} that across datasets, the affine transformations help separate the data in this $2$-dimensional space. The diffeomorphism-awareness of the DI $K$-means also drastically enhance the separability of the different clusters. When using DI $K$-means, the data are clustered based on macroscopically meaningful and interpretable parameters, making the model's performance possible to understand. 
For instance, for the Face-10 dataset, the t-SNE representation of the DI $K$-means clusters' shows that faces are grouped according to three significant orientations, left, right, and front. These three clusters are more easily observed in our DI $K$-means than in the affine invariant model. However, the $13$ different orientations present in the dataset remain too subtle to be captured by the DI $K$-means. These observations could explain the improved clustering performance of our model. 
%
%Interestingly, we can distinguish a fourth cluster that only contains faces of female subjects independent of their orientations. \randall{why is it interesting ? relate to euclidean distance or why kmean would not do that etc etc}
%
%It is nonetheless important to acknowledge that the performance of the DI $K$-means remains low compared to what supervised learning models would obtain, which are not competitors to our model. 
%
% For instance, for the Face-10 dataset, while the clustering accuracy is not high for any of the $K$-means methods, we observe in the t-SNE representation of the DI $K$-means case that the faces are grouped according to three significant orientations, left, right, and front, which indicates that the $13$ different orientations are too subtle to be characterized by the algorithm.
%
%
%Besides, on the upper right of the figure, we can see that there is a fourth cluster that contains the images of the only female belonging to the dataset.   

For the MNIST dataset, the last row and column of Fig.~\ref{fig:tsne}, we observe that most of the incorrectly ``classified" images are almost indistinguishable from samples of the other class. 
In particular, we highlight this by proposing to zoom-in into the cluster of $9$ in Fig.~\ref{fig:tsne}. We can see that the yellow instances are samples from the class $4$ that have been transformed such that they resemble the $9$'s centroid in Fig.~\ref{fig:centroids}. Besides, the centroid of the cluster $4$, in Fig.~\ref{fig:centroids}, has its upper part disconnected, while these transformed samples do not.
% 
%
%
%In the zoomed part of the $9$'s cluster, we observe the instances of $4$ that were transformed such that they resemble $9$ samples. 
%
%
%For the cluster of $1$, we observe transformed sampled of $5$ that have been narrowed as to become instances of $1$.
%
% For the MNIST dataset, the last row of Fig.~\ref{fig:tsne}, we observe that most of the instances that are not correctly ``classified" look like samples of the other cluster. 
%
We also provide a zoom-in on one of the clusters obtained on the rock-paper-scissors dataset, first row and last column of Fig.~\ref{fig:tsne}. The incorrectly clustered data are the ones that, when transformed, easily fit the scissor shape. In fact, we can see that both the transformed rock sample (yellow) and the paper (red) ones have the thumb pointing outward as the scissors centroid in Fig.~\ref{fig:centroids}.

The diffeomorphism-aware algorithm we propose is based on the transformation of any data into any centroid. Therefore, it depends on the number of landmarks, $\ell$, which can create transformations that are too large for some data and not enough for others. While such a trade-off is sample dependant, the overall results display a large improvement compared to the two other interpretable methods.

%For instance, the transformation required to transform a European $7$, that is, with a line in the middle, into a $7$ without a line can require a higher number of landmarks in the TPS method than the one that transform a $3$ into an $8$. 

\vspace{-.3cm}

\section{Conclusion}
We proposed a novel formulation of $K$-means suited to tackle clustering in computer vision tasks where image deformations should not affect the clustering of the algorithm. We did so by parametrizing this space of deformation with the TPS interpolation technique between images and centroids. 
Our technique reaches state-of-the-art performances while preserving the interpretability of the clustering as the centroids, and the TPS-transformed input images all live in the original data space. Interpretability of the model allows to easily diagnose failure cases, clustering decisions and should be favored in computer vision clustering application where explainability of the decision plays a role.
%
%Extension of this method for more complex images
%The update rules induced by the DI $K$-means similarity measure takes into account the transformations of the data and provide centroids that lie on the image manifold and can be fully interpreted as to understand the clustering decision of the algorithm.
%Finally, the experiments were performed in a controlled environment, where the images do not have any background. In fact, transforming objects in an image in the presence of background would drastically alter any transformation frameworks performed using an appearance-manifold  approach.

\section*{Acknowledgment}
This work was supported by
NSF grants SCH-1838873, CCF-1911094, IIS-1838177, and IIS-1730574;
NIH grant R01HL144683-CFDA;
ONR grants N00014-18-12571 and N00014-17-1-2551;
AFOSR grant FA9550-18-1-0478;
DARPA grant G001534-7500; and a
Vannevar Bush Faculty Fellowship, ONR grant N00014-18-1-2047.

{\small
\bibliographystyle{ieeetr}
\bibliography{egbib}
}

\clearpage
\onecolumn

\section{Properties of DI K-means and Proofs}
\label{app:proofs}

\subsection{DI K-means Similarity Measure: a Quasipseudosemimetric}

\begin{prop}
The similarity measure defined by $\min_{\nu \in \mathbb{R}^{2\ell}} \| \text{TPS}_{\ell}(x,\nu) - \mu \|$ is a Quasipseudosemimetric.
\end{prop}

\begin{proof}
Let's first define the orbit of an image with respect to the TPS transformations. Note that, the TPS does not form a group as it is a piecewise mapping. However, we know that it approximates the group of diffeomorphism on $\mathbb{R}^2$. Therefore, for sake of simplicity, we will make a slight notation abuse by considering the orbit, equivariance, and others group specific properties as being induced by the TPS.
\begin{defn}
\label{def:orbit_clust}
We define the orbit an image $x$ under the action the $\text{TPS}_{\ell}$ by
\begin{equation}
    \mathcal{O}(x) = \left \{\text{TPS}_{\ell}(x,\nu) | \nu \in \mathbb{R}^{2 \ell} \right \}.
\end{equation}
\end{defn}

Let's now consider each metric statement:
\\
\\

1) It is non-negative as per the use of a norm.

2) \textbf{Pseudo:} $\min_{\nu \in \mathbb{R}^{2\ell}} \left \| (\text{TPS}_{\ell}(x,\nu) - \mu \right \|=0 \Leftrightarrow \exists \nu \in \mathbb{R}^{2\ell}, s.t. \;\; x = \text{TPS}_{\ell}(x,\nu) \Leftrightarrow  x \sim_{\text{TPS}_{\ell}} \mu$, that is, $x$ and $\mu$ are equivariant with respect to the transformations induced by $\text{TPS}_{\ell}$. Thus, $d(x,\mu)=0$ for possibly distinct values $x$ and $\mu$, however, these are not distinct when we consider the data as any possible point on their orbit with respect to the group of diffeomorphism. In fact, the distance is equal to $0$ if and only if, $\mu$ and $x$ are equivariant. 

3) \textbf{Quasi:} The asymmetry of the distance is due to the non-volume preserving deformations considered. In fact, we do not consider the Haar measure of the associated diffeomorphism group and consider the $L_2$ distance with respect to the Lebesgue measure. Although the asymmetry of $d$ does not affect our algorithm or results, a symmetric metric can be built by normalizing the distance by the determinant of the Jacobian of the transformation. Such a normalization would make the metric volume-preserving and as a result make the distance symmetric. 

4) \textbf{Semi:} If $x,x',x'' \in \mathcal{O}$, then $d(x,x'')= d(x,x')= d(x',x'') = 0$ as it exist a $\nu, \nu', \nu''$ such that the TPS maps each data onto the other as per definition of the orbit, thus the triangular inequality holds. If $x,x'' \in \mathcal{O}$ and $x' \notin \mathcal{O}$, we have $d(x,x'')= 0 \leq d(x,x') + d(x',x'') $. If  $x,x' \in \mathcal{O}$ and $x
''\notin \mathcal{O}$, we have $d(x',x'') = d(x,x'')$, and since $0 \leq d(x,x')$, the inequality is respected. However, if $x,x',x''$ belong to three different orbits, then we do not have the  guarantee that then triangular inequality holds. In fact, it will depend on the distance between the orbits which is specific to each dataset. 
\end{proof}

\subsection{DI K-means Updates: Proof of Proposition~\ref{prop2}}
We consider the F\'echet mean of the centroid $k$ to be the solution of the following optimization problem, $\argmin_{\mu_k} \sum_{i: x_i \in C_k} d(x_i,\mu_k)$. Using our similarity measure, we obtain the following.
\begin{proof} 
\label{proof:prop2}
The Fr\'echet mean for the cluster $C_k$ is defined as $\argmin_{\mu_k, \left \| \mu_k \right \| =1 } \sum_{i: x_i \in C_k} \left \|  \text{TPS}_{\ell}(x_i,\nu^{\star}) - \mu_k  \right \|^{2}$
since the optimization problem is convex in $\mu_k$ (as the result of the composition of the identity map and a norm which are both convex) we have $\mu_k^{\star}: \nabla_{\mu} \sum_{i: x_i \in C_k} \left \| \text{TPS}_{\ell}(x_i,\nu^{\star}) - \mu_k  \right \|^{2} + \lambda (\left \| \mu_k \right \|-1) = 0$.
%Let's first find the differential,
%\begin{align}
%    \sum_{i: s \in C_k} & \left \|  \mathcal{T}(g,s) - (\mu_k + h)  \right \|^{2} -\sum_{i: s \in C_k} \left \|  \mathcal{T}(g,s) - %\mu_k   \right \|^{2} \nonumber \\
%     & = 2   \sum_{i: s \in C_k} \langle \mu_k + \mathcal{T}(g,s), h \rangle + o(  \left \|h  \right \|)
%\end{align}
%since $\mu_k$ belongs to the Hilbert space $\mathbb{R}^n$, the Riesz representation theorem implies that, 
with, 
\begin{align}
    \nabla_{\mu}  \sum_{i: x_i \in C_k} \left \|  \text{TPS}_{\ell}(x_i,\nu^{\star}) -  \mu_k  \right \|^{2} + \lambda \left \| \mu_k \right \|^2 & = 2 \: (\left | C_k \right | + \lambda) \times  \mu_k  + 2 \sum_{i: x_i \in C_k} \text{TPS}_{\ell}(x_i,\nu^{\star}).
\end{align}
%Note that $\sum_{i: x_i \in C_k} \text{TPS}_{\ell}(x_i,\nu^{\star}) \neq 0$ a.s.
\end{proof}

\subsection{DI K-means Similarity Measure: Invariance Property}
\label{app:invariance}
Motivated by the fact that small diffeomorphic transformations, usually, do not change nature of an image, we propose to exploit the invariance property of the similarity measure we proposed. We will also show that this invariance property induces generalization guarantees.

In this section, for sake of simplicity we will assume that the transformations belong to the group of diffeomorphism. In fact, the TPS can be considered as a method to sample element of this group.

Let's define an invariant similarity measure under the action of such group. That is, the similarity between two images remain the same under any diffeomorphic transformations. We propose to define the invariance in the framework of centroid-based clustering algorithm as follows.
\begin{defn}
\label{def:inv}
An invariant similarity measure with respect to $\text{Diff}(\mathbb{R}^2)$ is defined as $d: \mathbb{R}^n \times \mathbb{R}^n \rightarrow \mathbb{R}^{+}$ such that for all images $x \in \mathbb{R}^n$, all centroids $\mu \in \mathbb{R}^2$, and all group elements $\forall g \in \text{Diff}(\mathbb{R}^2)$, we have
\begin{equation}
\label{eq:inv}
d(x,\mu) = d( g \star x, \mu),
\end{equation}
where $g \star x$ denotes the action of the group element $g$ onto the image $x$.
\end{defn}

The similarity used in Eq.~\ref{eq:def_invariant} of the optimization problem is $\text{Diff}(\mathbb{R}^2)$-invariant as per Definition~\ref{def:inv}.
\begin{prop}
The similarity $ \min_{g \in \text{Diff}(\mathbb{R}^2)} \left \| g \star x - \mu \right \|$ is $\text{Diff}(\mathbb{R}^2)$-invariant.
\label{prop1}
\end{prop}
\begin{proof} 
\label{proof:prop}

Let consider $g^{\star} = \argmin_{g \in \text{Diff}(\mathbb{R}^2)} \left \| g \star x - \mu \right \|$, we have $\argmin_{g \in \text{Diff}(\mathbb{R}^2)} \left \| g \cdot g' \star x - \mu \right \| = g^{\star} \cdot g'^{-1}$, where $g'^{-1}$ is the inverse group element of $g'$. In fact, $\left \| g^{\star} \cdot g'^{-1} \cdot g' \star x - \mu \right \| = \left \| g^{\star} \star x - \mu \right \|.$ Since for all $g' \in \text{Diff}(\mathbb{R}^2)$, it exists an inverse element $g'^{-1}$, we have that $\forall g \in \text{Diff}(\mathbb{R}^2)$, $d( g' \star x, \mu) = d(x,\mu)$.

That is, by definition of the group, there is always another element that minimizes the loss function by using the composition between the inverse element of the group that has just been added, $g'$, and the optimal element $g^{\star}$.
\end{proof}

\section{Implementation Details}
\label{app:impl_details}
Note that the deformation invariant similarity measure we introduced in Eq.~\ref{eq:def_invariant} differs from the affine invariant distances developed in \cite{fitzgibbon2002affine,simard2012transformation,lim2004image}. All previously defined measures of error rely on the assumption that the manifold can be locally linearized and as a result the tangent space is used as a proxy to learn the optimal affine transformation. 
However, the work of \cite{wakin2005multiscale} suggests that tangent planes fitted to image manifold continually twist off into new dimensions as the parameters of the affine transformations vary due to a possible the intrinsic multiscale structure of the manifold. As such, the alignment of two images can be done by linearizing the manifold. To do so, \cite{wakin2005multiscale} propose to consider the multiscale structure of the manifold, we simplify their approach by applying a low-pass filter on the images and the centroid prior to learn the affine transformation best aligning them. Then, we optimize the remaining part of the TPS to account for diffeomorphic transformations. These two steps are similar to the one used in \cite{jaderberg2015spatial}.

\section{Neural Network Architectures}
\label{app:arch}

For both architectures , the decoder architecture is symmetric to the encoder and the batch size is set to $64$. 
\paragraph{MLP:} The MLP architecture from input data to bottleneck hidden layer is composed of $4$ fully connected ReLU layers with dimensions $\left [500,500,2000,10 \right ]$.

\paragraph{Conv:} The CONV architecture is composed of $3$ $2d$-convolutional ReLU layers with $32$ filters of size $5\times5$, and $2$ fully connected ReLU layers with dimension $\left [400, 10 \right ]$. For each layer, a batch normalization is applied.

\section{Thin-Plate-Spline Interpolation}
\label{app:TPS}
Let's consider two set of landmarks, the source ones $\nu_s = \{u_i,v_i\}_{i=1}^\ell$ and the transformed $\nu_t = \{u_i',v_i'\}_{i=1}^\ell$ where $\ell$ denotes the number of landmarks. The TPS aim at finding a mapping $F=(F_1,F_2)$, such that $F(u,v) = (F_1(u,v),F_2(u,v)) = (u',v')$, that is, the mapping between two set of landmarks. The particularity of the TPS is that it learns such a mapping by minimizing the interpolation term, and a regularization that consists in penalizing the bending energy. 

The TPS optimization problem is defined by
\begin{equation}
    \min_{F} \sum_{i=1}^{N} \left \|(u_i',v_i')-F(u_i,v_i)  \right \|^2 + \lambda \int \int \left [ (\frac{\partial ^2 F}{\partial u^2})^2 + 2 (\frac{\partial^2 F}{\partial u \partial v})^2 + (\frac{\partial ^2 F}{\partial v^2})^2  \right ] du dv.
\label{eq:bend}
\end{equation}

In our model, the source landmarks are consider to be the coordinates of a uniform grid. Also note that both the source landmarks and transformed ones are usually a subsets of the set of coordinate of the images. For instance, for the MNIST dataset of size $28\times28$, the landmarks would be a grid of size $\ell \times \ell$, where $\ell <28$. While the mapping is based on the landmark, it is then applied to the entire image coordinate. In fact, $F=(F_1,F_2)$ is mapping $\mathbb{R}^2 \rightarrow \mathbb{R}^2$, where  $F_1$ (resp. $F_2$) corresponds to the mapping from $(x,y)$ to the first dimension $x'$ (resp. the second dimension $y'$).

The solution of the TPS optimization problem, Eq.~\ref{eq:bend}, provides the following analytical formula for $F$
\begin{align}
\label{eq:tpsx}
     F_1(u,v) = \hspace{-.1cm} u'   \hspace{-.1cm} =   \hspace{-.1cm} a_1^{(1)} &  \hspace{-.1cm} + \hspace{-.1cm} a_{u}^{(1)}u \hspace{-.1cm}  + \hspace{-.1cm}  a_{v}^{(1)} v \hspace{-.1cm}  + \hspace{-.1cm} \sum_{i=1}^{\ell} \hspace{-.1cm}  w_i^{(u)}   U(\left | (u_i,v_i)\hspace{-.1cm}  -\hspace{-.1cm}  (u,v) \right |),
\end{align}
\begin{align}
\label{eq:tpsy}
     F_2(u,v) = \hspace{-.1cm} v'  \hspace{-.1cm} =   \hspace{-.1cm} a_1^{(2)} & \hspace{-.1cm}  + \hspace{-.1cm} a_{u}^{(2)}u \hspace{-.1cm} + \hspace{-.1cm} a_{v}^{(2)} v \hspace{-.1cm}  +\hspace{-.1cm}  \sum_{i=1}^{\ell} \hspace{-.1cm}  w_i^{(v)}  U(\left | (u_i,v_i) \hspace{-.1cm} - \hspace{-.1cm} (u,v) \right |),
\end{align}
where $\left | . \right |$ is the $L_1$-norm, $a_1,a_{u},a_{v}$ are the parameters governing the affine transformation, and $w_i$ are parameters responsible for non-rigid transformations as they stand as a weight of the non-linear kernel $U$. The non-linear kernel $U$ is expressed by $U(r) = r^2 \log(r^2), \forall r \in \mathbb{R}_{+}$.

Based on the landmarks $\nu_s$ and $\nu_t$, we can obtain these parameters by solving a simple system of equation define by the following operations

\begin{equation}
\label{eq:solving}
    \mathcal{L}^{-1}\mathcal{V} = \begin{bmatrix}
(W^{(x)}|a_1^{(x)} a_x^{(x)} a_y^{(x)})^T\\ 
(W^{(x)}|a_1^{(y)} a_x^{(y)} a_y^{(y)})^T
\end{bmatrix}.
\end{equation}
where the matrix $\mathcal{L} \in \mathbb{R}^{(\ell+3) \times (\ell+3)}$, is defined as
\[ \mathcal{L} =
\left[
\begin{array}{c|c}
\mathcal{K} & \mathcal{P} \\
\hline
\mathcal{P}^T & \mathcal{O}
\end{array}
\right],
\mathcal{K} = \begin{bmatrix}
 0  & U(r_{12}) & \dots & U(r_{1\ell}) \\ 
 U(r_{21}) & 0 & \dots & U(r_{2\ell}) \\ 
 \dots & \dots & \dots & \dots \\ 
 U(r_{\ell1}) & \dots & \dots & 0 
\end{bmatrix}, 
    \mathcal{P} = \begin{bmatrix}
 1  & x_1  & y_1 \\ 
 1 &  x_2 & y_2 \\ 
 \dots & \dots & \dots  \\ 
 1 & x_{\ell} & y_{\ell} 
\end{bmatrix} \]
where $r_{ij}= \left | (u_i,v_i) - (u_j,v_j) \right |$, $\mathcal{K} \in \mathbb{R}_{+}^{\ell \times \ell}$,
and $\mathcal{V}=\begin{bmatrix}
x_1' & x_2' & \dots & x_\ell'| 0 &0 &0\\ 
y_1' & y_2' & \dots & y_\ell'| 0 &0 &0
\end{bmatrix}.$

Note that, since the matrix $\mathcal{L}$ depends only on the source landmarks, and that in our case these are unchanged, its inverse can be computed only once. The only operation required to be computed for each data and each centroid is the matrix multiplication $\mathcal{L}^{-1} \mathcal{V}$ providing the parameters of the TPS transformation, as per Eq.~\ref{eq:tpsx}, \ref{eq:tpsy}. Given these parameters, the mapping $F$ can be applied to to each coordinate of the image.

 Now in order to render the image, one can perform bilinear interpolation as it is achieved in . Besides, the bilinear interpolation will allow the propagation of the gradient through any differentiable loss function.

Given an image $x_1 \in \mathbb{R}^{n}$  where $n=W \times H$, $W$ denotes the width and $H$ the height of the image, and two sets of landmarks $\nu_s = \{u_i,v_i\}_{i=1}^\ell$ ,uniform grid coordinate of $x_1$, and $\nu_t = \{u_i',v_i'\}_{i=1}^\ell$, the transformation of the uniform grid, which are subset of the image coordinate,
We are able to learn a mapping $F=(F_1,F_2)$ such that for each original pixel coordinate, we have their transformed coordinates. In fact, given any position $(u,v)$ on the original image, the mapping $F$ provides the new positions $(u',v')$ as per Eq.~\ref{eq:tpsx}, Eq.~\ref{eq:tpsy}.

Now, from this transformed the coordinates space, we can render an image $x_2 \in \mathbb{R}^{n}$ using, as in \cite{jaderberg2015spatial}, the bilinear interpolation function $\Gamma: \mathbb{R}^2 \times \mathbb{R}^n \rightarrow \mathbb{R}$ which takes as input the original image $x_1$ and the transformed pixel coordinates $(u',v')$, and outputs the pixel value of the transformed image at a given pixel coordinate 
 \begin{equation*}
 \begin{aligned}
     x_2  (k,l)= & \Gamma[F(u_k,v_l),x_1] \nonumber \\
      = & \Gamma[(u_k',v_l'),x_1] \nonumber \\
      = & \sum_{t,h \in \{0,1\}}\sum_{i=1}^{W} \sum_{j=1}^{H} x_1(i,j) \delta(\floor{u_k'+t}-i) 
       \times \delta(\floor{v_l'+h}-j)(u_k'-\floor{u_k'})^{\delta(t)}(v_l'-\floor{v_l'})^{\delta(h)} \nonumber \\
       & \hspace{1cm} \times (1-(v_l'-\floor{v_l'}))^{\delta(t-1)}(1-(u_k'-\floor{u_k'}))^{\delta(h-1)}, 
     \end{aligned}
 \end{equation*}
 where $\delta$ is the Kronecker delta function and $\floor{.}$ is the floor function rounding the real coordinate to the closest pixel coordinate. 
 
%\begin{figure}[!h]

\section{Datasets}
\label{ap:data}
\noindent
\textbf{MNIST \cite{deng2012mnist}:} is a handwritten digit dataset containing $60.000$ training and  $10.000$ test images of dimension $28 \times 28$ representing $10$ classes.
\\
\noindent
\textbf{Aff. MNIST:} we randomly sample one instance of each MNIST class and generate $100$ random affine transformations for each sample.  A third of the data are used for testing.
\\
\noindent
\textbf{Diffeo. MNIST:} we randomly sample one instance of each MNIST class and generate $100$ random affine transformations for each sample as well as random diffeomorphisms using the TPS method. A third of the data are used for testing.
\\
\noindent
\textbf{Audio MNIST \cite{becker2018interpreting}:} is composed of $30000$ recordings of spoken digits by  $60$  different speaker and sampled at $48kHz$ of $1$sec long. It consists of $10$ classes. We use $10000$ data for testing and $20000$ for testing. This dataset will be transformed into a time-frequency representation. This representations, can be considered as images, are usually used as the common representation of audio recordings \cite{cosentino2020}. 
\\
\noindent
\textbf{E-MNIST \cite{cohen2017emnist}:} is a handwritten letters dataset merging a balanced set of the uppercase and lowercase letters into a single $26$ classes dataset of dimension $28 \times 28$.
\\
\noindent
\textbf{Rock-Paper-Scissors \cite{rps}:} images of hands playing rock, paper, scissor game, that is, a $3$ classes dataset. The dimension of each image is $300 \times 300$. The training set is composed of $2520$ data and the testing set $372$.
\\
\noindent
\textbf{Face-10 \cite{gourier2004estimating}:} images of the face of 15 people, wearing glasses or not and having various skin color. For each individual, different samples are obtained with different pose orientation varying from $-90$ degrees to $+90$ degrees vertical degrees. The dimension of each image is $288 \times 384$. The training set is composed of $273$ data and testing set of $117$ data.
\\
\noindent
\textbf{Arabic Char \cite{altwaijry2020arabic}:} Handwritten Arabic characters written by $60$ participants. The dataset is composed of $13,440$ images in the training set and $3360$ in the test set. The dimension of each image is $32 \times 32$.

\section{Supplementary Visualisations}

\subsection{Additional t-SNE Visualisations}
\label{app:sup_tsne}

\begin{figure*}[h!]

\begin{minipage}{.32\linewidth}
    \centering
\textbf{\textit{Raw Data}}
\end{minipage}
\begin{minipage}{.32\linewidth}
    \centering
\textbf{\textit{Affine Invariant}}
\end{minipage}
\begin{minipage}{.32\linewidth}
    \centering
\textbf{\textit{Deformation Invariant}}
\end{minipage}

\begin{minipage}{.01\linewidth}
\rotatebox{90}{\textbf{\textit{E-MNIST}}}
\end{minipage}
\begin{minipage}{.32\linewidth}
    \centering
    \includegraphics[width=\linewidth]{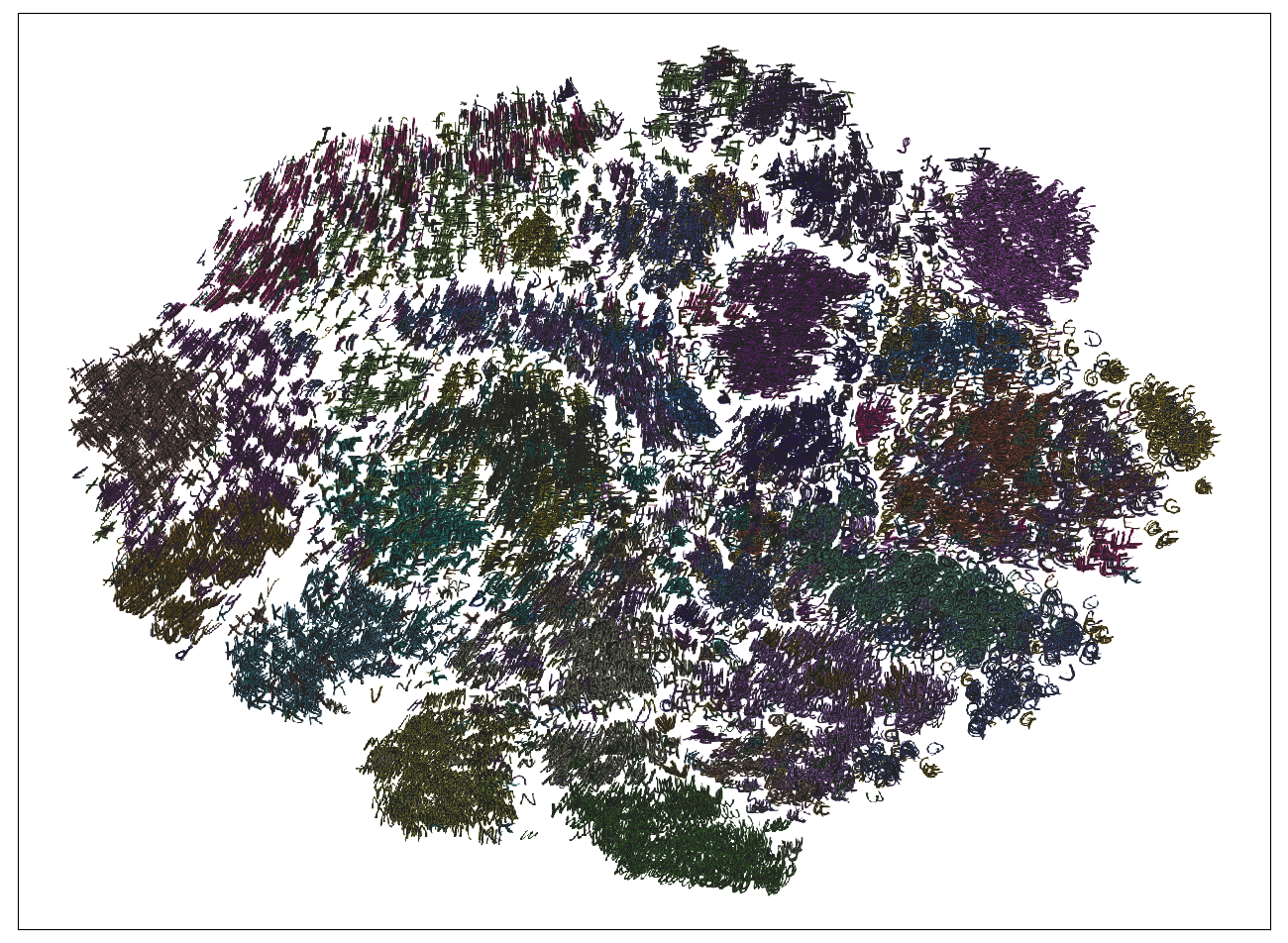}
\end{minipage}
\begin{minipage}{.32\linewidth}
    \centering
    \includegraphics[width=\linewidth]{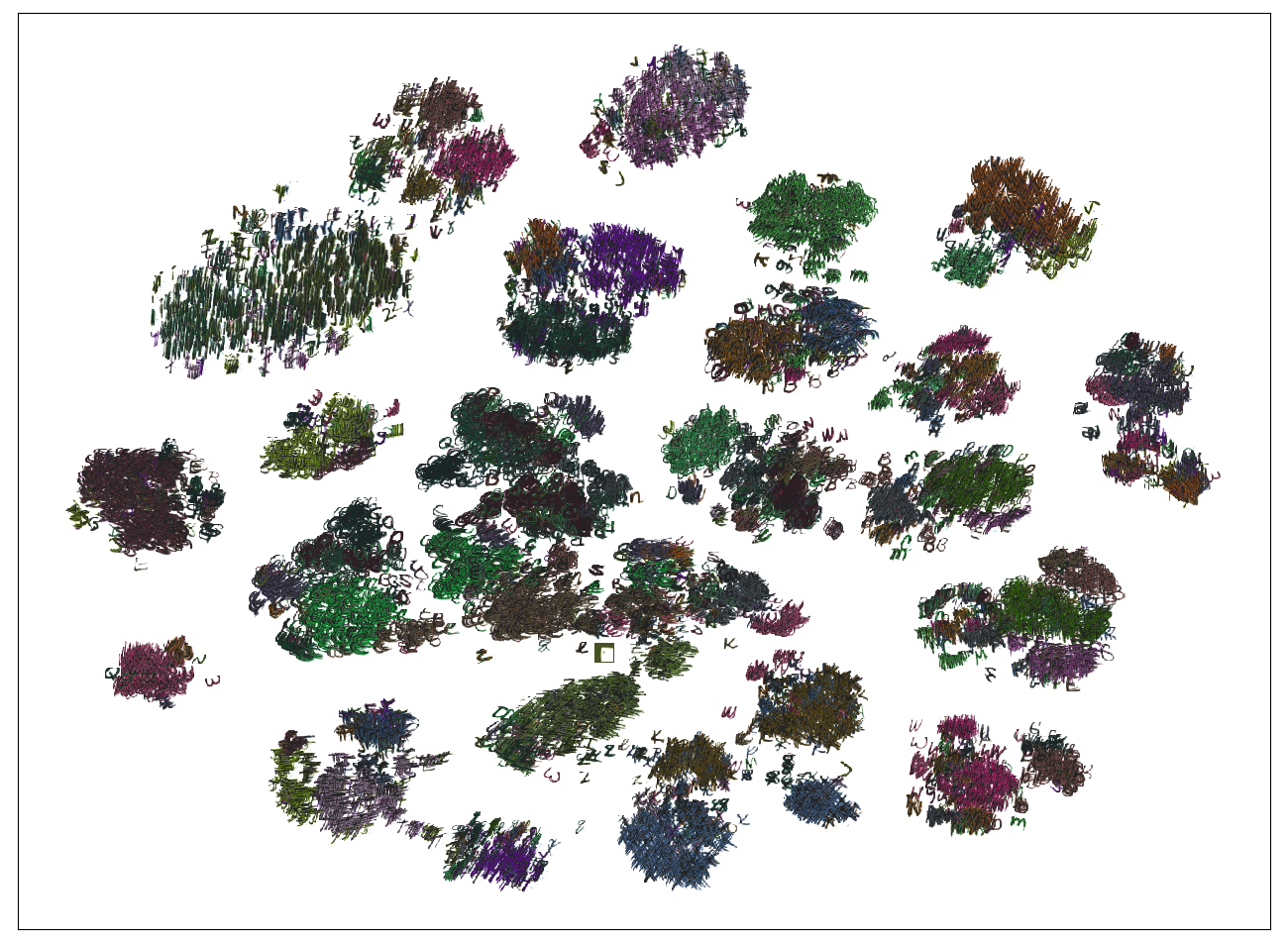}
\end{minipage}
\begin{minipage}{.32\linewidth}
    \centering
    \includegraphics[width=\linewidth]{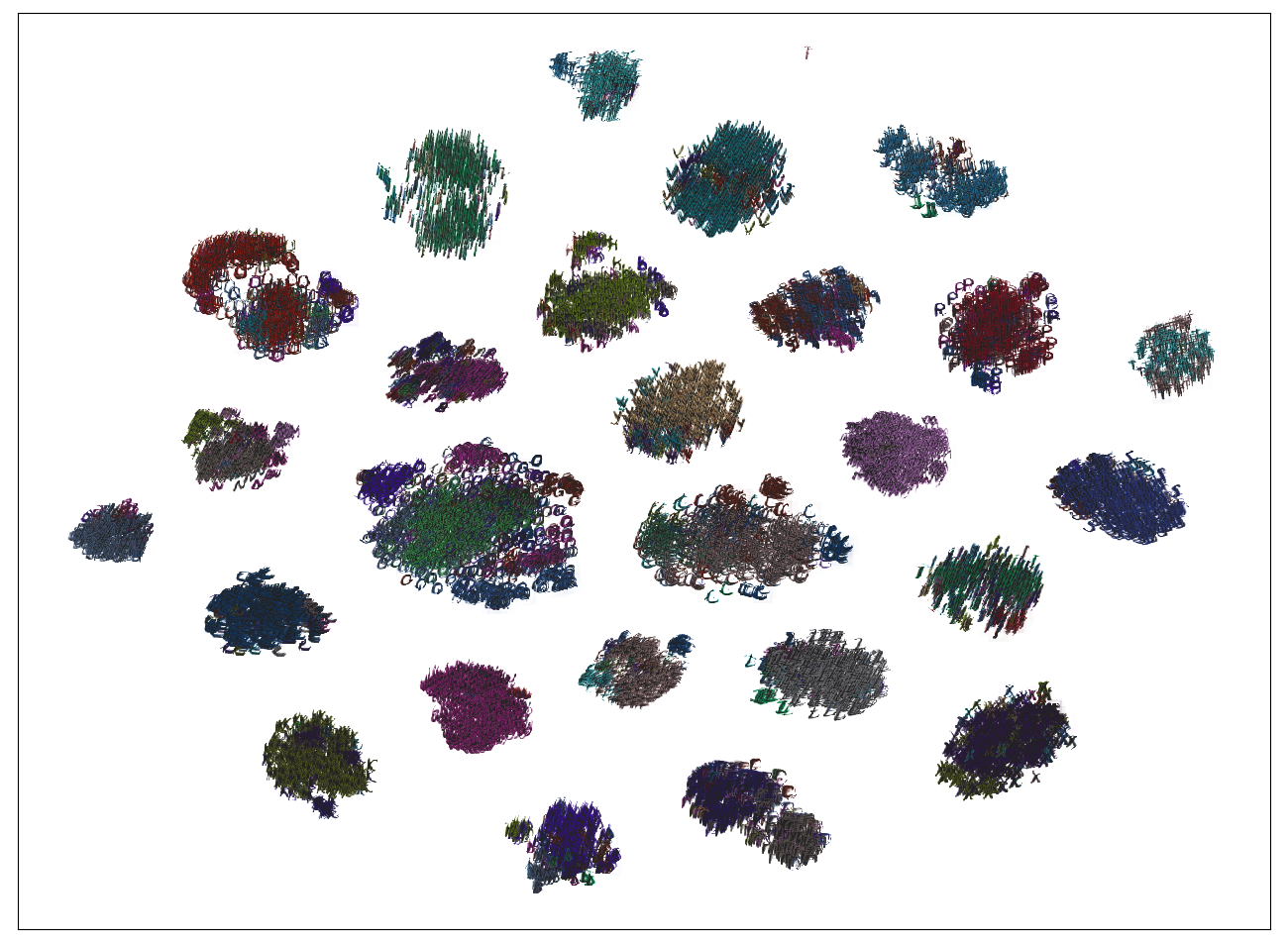}
\end{minipage}

\begin{minipage}{.01\linewidth}
\rotatebox{90}{\textbf{\textit{Arabic Characters}}}
\end{minipage}
\begin{minipage}{.32\linewidth}
    \centering
    \includegraphics[width=\linewidth]{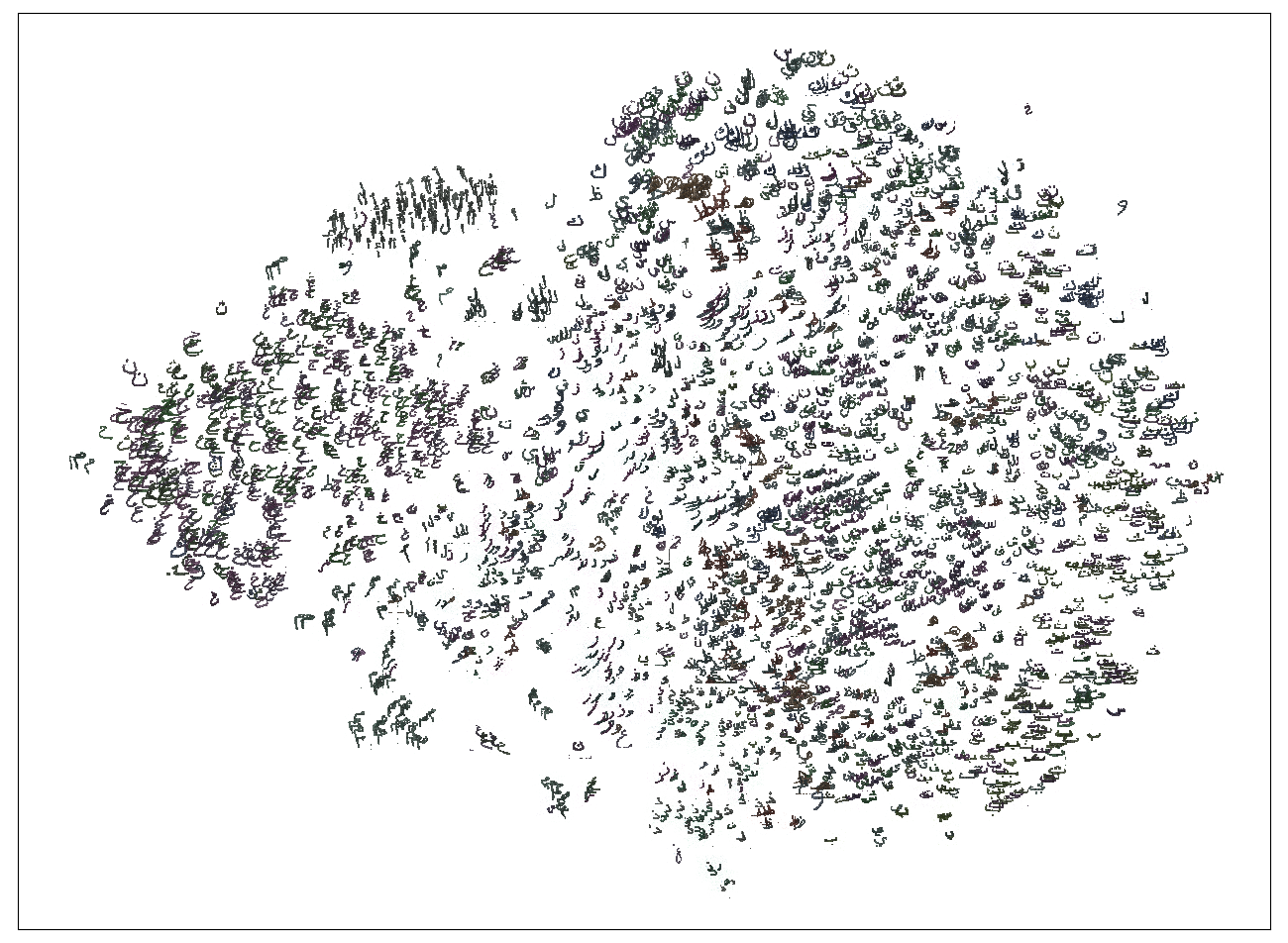}
\end{minipage}
\begin{minipage}{.32\linewidth}
    \centering
    \includegraphics[width=\linewidth]{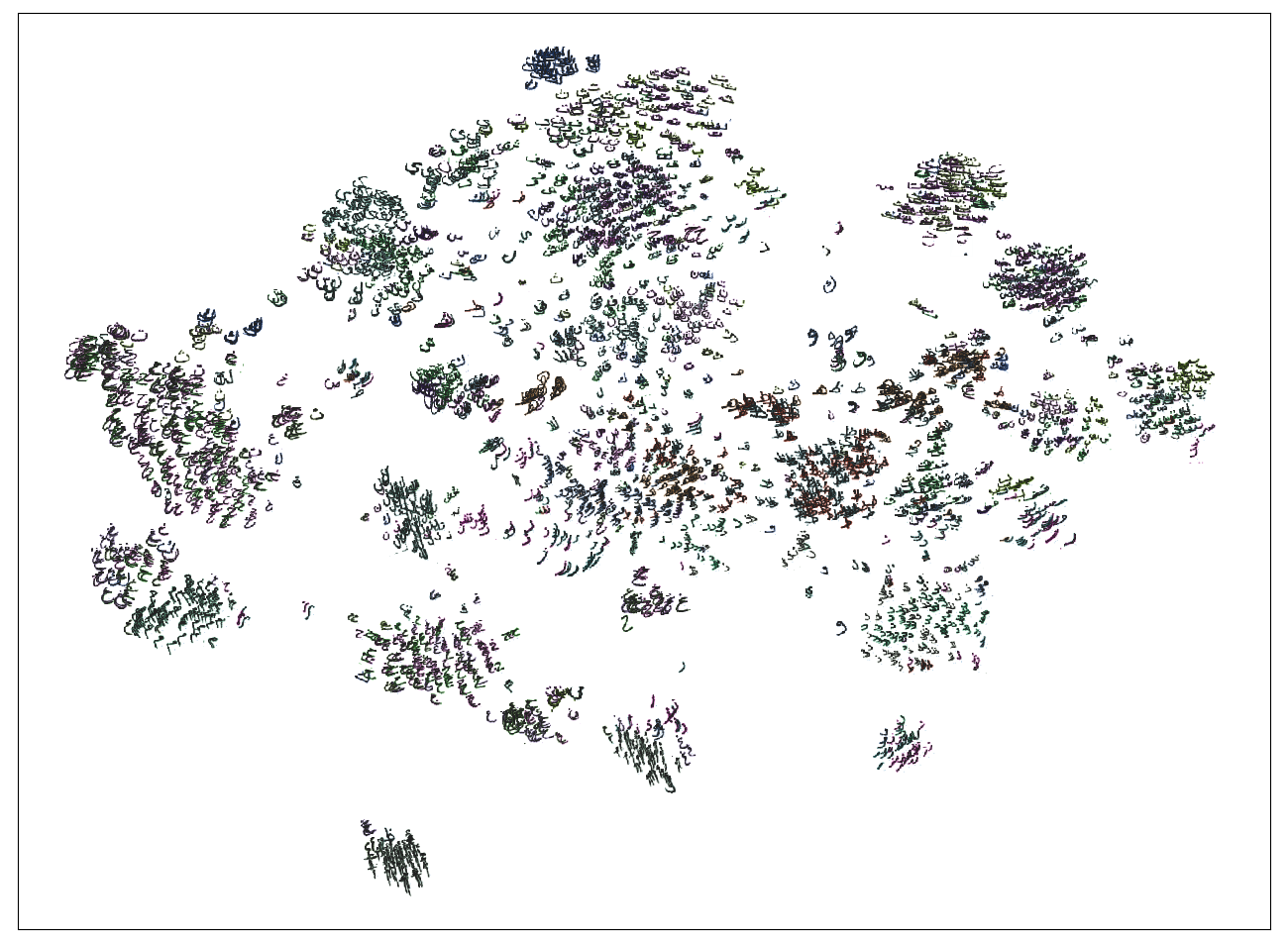}
\end{minipage}
\begin{minipage}{.32\linewidth}
    \centering
    \includegraphics[width=\linewidth]{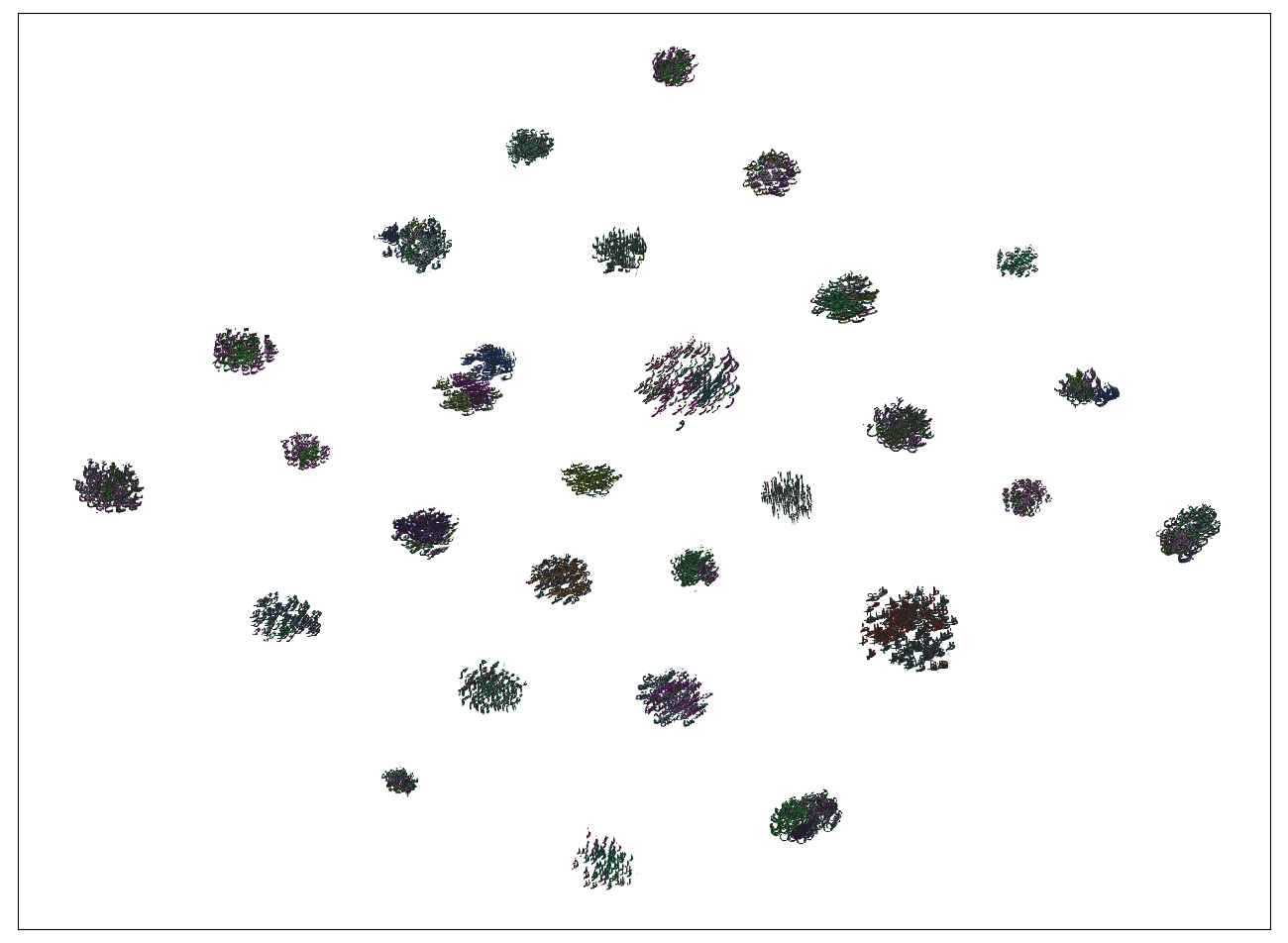}
\end{minipage}

\caption{\textbf{$2$-dimensional t-SNE Vizualisation} - The raw data (\textit{left column}), the affinely transformed data, i.e., we extract the transformation of the data that corresponds to the centroid it was assigned and perform the t-SNE on these affinely transformed data, (\textit{middle column}), the data transformed with respect to  diffeomorphism as per Eq.~\ref{eq:def_invariant}, i.e., the same process as previously mentioned but we consider the transformation induced by the TPS, and then perform the dimension reduction on these transformed data, (\textit{right column}). Each row corresponds to a different datasets, E-MNIST, Arabic Characters,  are depicted from the top to bottom row. For all the figures, the colors of the data represent their ground truth labels.}
\end{figure*}

\subsection{Additional Centroid Visualisations}
\label{app:centroids}

\begin{figure*}[h!]
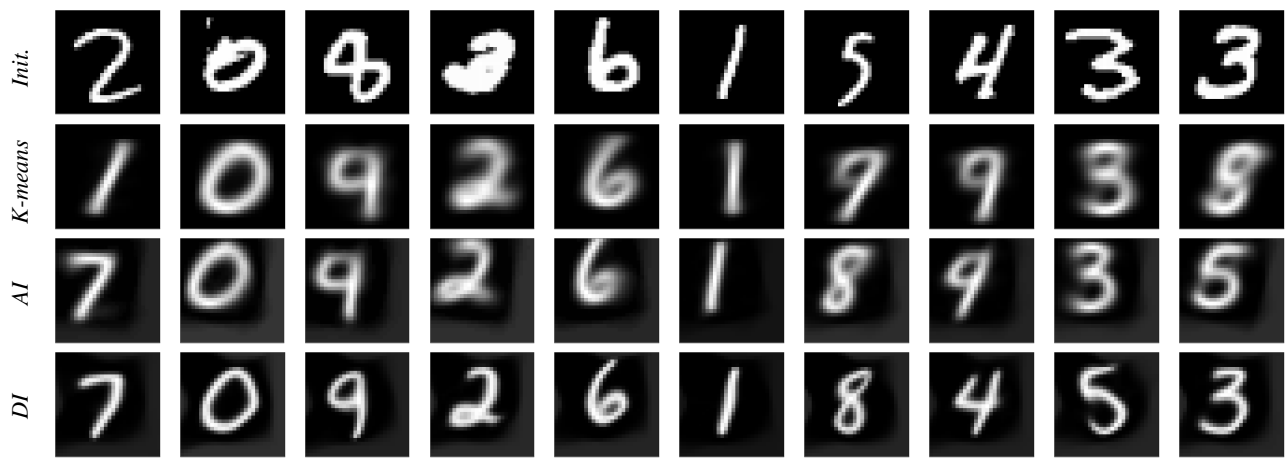

\begin{minipage}{.02\linewidth}
\rotatebox{90}{\small \textit{DI  \; \; \; \; \;\;\;\;\;AI \;\;\;\;\;\;\; K-means\;\;\;\;\;\;\;\;Init.}}
\end{minipage}
\foreach \c in {0,...,9}{
    \begin{minipage}{0.085\linewidth}
    \includegraphics[width=\linewidth]{images/best_mnist_centroid_init\c.png}\\
    \includegraphics[width=\linewidth]{images/kmean_mnist_centroid_final\c.png}\\
    \includegraphics[width=\linewidth]{images/aff_mnist_centroid_final\c.png}\\
    \includegraphics[width=\linewidth]{images/best_mnist_centroid_final\c.png}
    \end{minipage}
}
\caption{ \textit{Additional} - \textbf{MNIST Centroids Visualization (dim 28x28)} - Depiction of the initialization of the per-cluster centroids \textbf{top row} and the final per-cluster centroids of the K-means, Affine invariant K-means, and DI K-means (proposed) methods.}
\end{figure*}

\begin{figure*}[h!]
\begin{minipage}{.02\linewidth}
\rotatebox{90}{\small \textit{DI  \; \; \; \; \;\;\;\;\;AI \;\;\;\;\;\;\; K-means\;\;\;\;\;\;\;\;Init.}}
\end{minipage}
\foreach \c in {10,...,19}{
    \begin{minipage}{0.085\linewidth}
    \includegraphics[width=\linewidth]{images/best_emnist_centroid_init\c.png}\\
    \includegraphics[width=\linewidth]{images/kmean_emnist_centroid_final\c.png}\\
    \includegraphics[width=\linewidth]{images/aff_emnist_centroid_final\c.png}\\
    \includegraphics[width=\linewidth]{images/best_emnist_centroid_final\c.png}
    \end{minipage}
}
\caption{ \textit{Additional} - \textbf{$10$ out of $26$ EMNIST Centroids Visualization (dim 28x28)} - Depiction of the initialization of the per-cluster centroids \textbf{top row} and the final per-cluster centroids of the K-means, AI K-means, and DI K-means (proposed) methods.}
\end{figure*}

\begin{figure*}[h!]
\begin{minipage}{.02\linewidth}
\rotatebox{90}{\small \textit{DI  \; \; \; \; \;\;\;\;\;AI \;\;\;\;\;\;\; K-means\;\;\;\;\;\;\;\;Init.}}
\end{minipage}
\foreach \c in {0,...,9}{
    \begin{minipage}{0.085\linewidth}
    \includegraphics[width=\linewidth]{images/best_audiomnist_centroid_init\c.png}\\
    \includegraphics[width=\linewidth]{images/kmean_audiomnist_centroid_final\c.png}\\
    \includegraphics[width=\linewidth]{images/aff_audiomnist_centroid_final\c.png}\\
    \includegraphics[width=\linewidth]{images/best_audiomnist_centroid_final\c.png}
    \end{minipage}
}
\caption{ \textit{Additional} - \textbf{Audio MNIST Centroids Visualization (dim 64x24)} - Depiction of the initialization of the per-cluster centroids \textbf{top row} and the final per-cluster centroids of the K-means, AI K-means, and DI K-means (proposed) methods.}
\end{figure*}

\begin{figure*}[h!]
\begin{minipage}{.02\linewidth}
\rotatebox{90}{\small \textit{DI  \; \; \; \; \;\;\;\;\;AI \;\;\;\;\;\;\; K-means\;\;\;\;\;\;\;\;Init.}}
\end{minipage}
\foreach \c in {0,...,9}{
    \begin{minipage}{0.085\linewidth}
    \includegraphics[width=\linewidth]{images/best_arabchar_centroid_init\c.png}\\
    \includegraphics[width=\linewidth]{images/kmean_arabchar_centroid_final\c.png}\\
    \includegraphics[width=\linewidth]{images/aff_arabchar_centroid_final\c.png}\\
    \includegraphics[width=\linewidth]{images/best_arabchar_centroid_final\c.png}
    \end{minipage}
}
\caption{ \textit{Additional} - \textbf{$10$ out of $28$ Arab Characters Centroids Visualization (dim 32x32)} - Depiction of the initialization of the per-cluster centroids \textbf{top row} and the final per-cluster centroids of the K-means, AI K-means, and DI K-means (proposed) methods.}
\end{figure*}

%
%(\textit{\nth{2} Row}) Centroids discovered by the K-means algorithm. (\textit{\nth{3} Row}) Centroids learned by optimizing only the affine transformation of leading to invariance with respect to affine transformation only. (\textit{\nth{4} Row}) The final centroids learned through the Fr\'echet mean described in Proposition~\ref{prop2}. The update given our metric builds centroids as an average of transformed digits. As opposed to Euclidean distance based update, there is no overlapping of different instances of the same cluster producing centroids composed of a mixture of different images, which, depending on the intra-class variance can lead to erroneous cluster assignment.}

\begin{figure*}[h!]
\begin{minipage}{.02\linewidth}
\rotatebox{90}{\small \textit{DI  \; \; \; \; \;\;\;\;\;AI \;\;\;\;\;\;\; K-means\;\;\;\;\;\;\;\;Init.}}
\end{minipage}
\foreach \c in {3,...,12}{
    \begin{minipage}{0.085\linewidth}
    \includegraphics[width=\linewidth]{images/best_facepos_centroid_init\c.png}\\
    \includegraphics[width=\linewidth]{images/kmean_facepos_centroid_final\c.png}\\
    \includegraphics[width=\linewidth]{images/aff_facepos_centroid_final\c.png}\\
    \includegraphics[width=\linewidth]{images/best_facepos_centroid_final\c.png}
    \end{minipage}
}
\caption{ \textit{Additional} - \textbf{$10$ out of $13$ Face Position Centroids Visualization (dim  288x384)} - Depiction of the initialization of the per-cluster centroids \textbf{top row} and the final per-cluster centroids of the K-means, AI K-means, and DI K-means (proposed) methods.}
\end{figure*}

\end{document}